\documentclass[a4paper]{article}

\usepackage{amsmath}
\usepackage{amsthm}
\usepackage{rotating} % for sidewaystable
\usepackage{longtable}

\usepackage{amssymb}
\usepackage{mathtools}
\usepackage{graphicx}
\usepackage[margin=1in]{geometry}
\usepackage[colorlinks=true, allcolors=blue]{hyperref}
\usepackage{xparse}
\usepackage{todonotes}
\usepackage{xspace}
\usepackage{tikz}
\usepackage{xparse}
\usetikzlibrary{overlay-beamer-styles}
\usepackage{subcaption}
\usepackage{etoolbox}
\usepackage[capitalize,noabbrev]{cleveref}
\usepackage{siunitx}
\usepackage{booktabs}
\usepackage{algorithm}
\usepackage{algpseudocode}

\usepackage{authblk}
\providecommand{\keywords}[1]{\small\textbf{\textit{Keywords---}} #1}

\presetkeys{todonotes}{inline}{}

\DeclareDocumentCommand\R{}{\mathbb{R}}
\DeclareDocumentCommand\Rnonneg{}{\R_{\geq 0}}
\DeclareDocumentCommand\Z{}{\mathbb{Z}}
\DeclareDocumentCommand\Znonneg{}{\Z_{\geq 0}}
\DeclareDocumentCommand\orderO{o}{\ensuremath{\mathcal{O}\IfValueTF{#1}{\left(#1\right)}{}}}

\title{A hybrid solution approach for the Integrated Healthcare Timetabling Competition 2024}

\author[1,2]{Daniela Guericke\thanks{d.guericke@utwente.nl}}
\author[3]{Rolf van der Hulst\thanks{rolf.vanderhulst@mosek.com (now affiliated with Mosek ApS)}}
\author[1,2]{Asal Karimpour\thanks{a.karimpour@utwente.nl}}
\author[1,2]{Ieke Schrader\thanks{i.m.w.schrader@utwente.nl}}
\author[3]{Matthias Walter\thanks{m.walter@utwente.nl}}

\affil[1]{University of Twente, Center for Healthcare Operations Improvement and Research, Enschede, The Netherlands}
\affil[2]{University of Twente, Industrial Engineering and Management Science, Enschede, The Netherlands}
\affil[3]{University of Twente, Mathematics of Operations Research, Enschede, The Netherlands}

\DeclareDocumentCommand\edit{m}{\textcolor{red}{#1}}
\DeclareDocumentCommand\edit{m}{#1}

\begin{document}

\maketitle

\begin{abstract}
  In this work, we present the solution approach for the Integrated Healthcare Timetabling Competition 2024 submitted by Team Twente, which ultimately ranked third among the finalists.
  Our approach combines mixed-integer programming, constraint programming, and simulated annealing in a 3-phase solution approach based on decomposition into subproblems.
  In addition to describing our approach and design decisions, we share our insights and, for the first time, lower bounds on the optimal solution values for the benchmark instances.
  We analyze the results based on solution quality \edit{for the competition and an extended runtime Additionally, we investigate the different soft constraints} and specific parts of the algorithm.
  Finally, we highlight open problems and future research directions \edit{for further improving} the approach.
\end{abstract}

\keywords{%
  integrated planning, nurse assignment, patient admission, healthcare, room assignment
}%

\section{Introduction}
\label{sec_intro}

Our healthcare systems are struggling with the ageing population, resulting in an increasing demand and rising expenditures, while simultaneously facing a shortage of healthcare professionals~\cite{OECD2024, Zapata2023}.
When a system is under stress and demand exceeds supply, among other strategies, efficient resource planning and scheduling become important~\cite{Page2024}.
Hospitals are a critical component of the healthcare system, playing a vital role in care coordination, system development, and supporting population health needs~\cite{WHO_Hospitals}.

Efficient planning and scheduling in hospitals is important to utilize the limited resources in the best possible manner.
Here, approaches from Operations Research can be of benefit to optimize planning problems such as admission planning, room allocation, nurse assignment, and surgery scheduling~\cite{Hulshof2012}.
It has been recognized that resources should be planned in an integrated manner to improve the overall outcomes instead of focusing on individual departments or resources~\cite{rachuba2024integrated}.
However, \edit{integrating} several decision problems into one increases complexity, \edit{therefore} calling for new solution approaches.
Based on these observations, the \emph{Integrated Healthcare Timetabling Competition} (IHTC) was launched in~2024, \edit{which focuses} specifically on finding efficient solution approaches for the integrated planning of patient admissions, room allocation, nurse assignment, and operating theater (OT) planning~\cite{IHTC24}.
A detailed description of the competition is given in~\cite{IHTC24}.

In this paper, we describe our hybrid solution approach to the IHTC~2024.
Since the planning problem \edit{proves} to be computationally challenging to solve in an integrated manner already for small instances, we propose a 3-phase decomposition approach that combines Mixed-Integer Linear Programming (MIP), Constraint Programming (CP), and Simulated Annealing (SA).
This method earned a finalist position and ultimately ranked third place. 

The remainder of this paper is organized as follows.
\Cref{sec_literature} presents related work on integrated planning in hospitals.
\Cref{sec_problemsetting} gives an overview of the underlying planning problem, including subproblems, decisions, and constraints.
We refer the reader to~\cite{IHTC24} for the original problem statement.
\Cref{sec_approach} describes the three phases of our decomposition approach and presents details for all subproblems addressed by the approach.
\Cref{sec_rejection} presents a variant of the solution approach that we developed after the competition.
We report on our computational results in \cref{sec_experiments} and conclude the paper with observations and ideas for future research in \cref{sec_conclusions}.

\section{Related work}
\label{sec_literature}

Traditionally, healthcare planning aimed to improve efficiency by managing and allocating resources separately within individual hospital departments~\cite{rachuba2024integrated}.
However, these isolated planning procedures often overlook the interactions and interdependencies among departments~\cite{Hulshof2012,rachuba2024integrated}.
Consequently, an increasing need for integrated resource planning approaches has emerged, in which multiple resources and decision levels are jointly considered within healthcare organizations~\cite{leeftink2020multi}.
In this section, we provide a literature overview of integrated planning \edit{problems that consider resources similar to those} in the IHTC, \edit{as well as} frequently used approaches for those problems.

Rachuba, Reuter-Oppermann and Thielen~\cite{rachuba2024integrated} \edit{classify} planning approaches in hospitals based on the extent to which multiple resources are integrated in one planning problem.
A level-3 integrated approach is the highest level of integration, in which all resource decisions are taken simultaneously and presented in a single solution.
In a level-2 integration, resource planning problems are solved sequentially, with a solution to an earlier problem used as input to a later one.
On the lowest level, a level-1 integration, resources are independently planned but restricted by other resources. The restrictions are independent of the solution \edit{to} the planning problems of the other resources.
In our solution approach, we decompose the overall problem into several subproblems.
We sequentially decide upon patient admissions, patient-to-room assignment, patient-to-OT assignment and nurse-to-room assignment.
We use earlier made decisions in the later subproblems.
Moreover, we incorporate lower bounds from later subproblems into earlier ones.
Therefore, we classify our developed approach as a level-2 integrated planning approach.

Analysis of the level-2 and level-3 integrations in the supplementary data from~\cite{rachuba2024integrated} shows that, among the resources in the IHTC, OTs are the most frequently planned primary resource (117 occurrences), followed by nurses (65 occurrences) and patient admissions (45 occurrences).
Primary resources are decision variables within the planning problem.
In contrast, rooms are rarely a primary resource, appearing only eight times in sequential (level 2) and simultaneous planning (level 3). 
Consequently, the combination of the four resources in the IHTC remains scarce in the literature.

This scarcity is even more pronounced when narrowing the focus to level-2 and level-3 integrations at the strictly operational level, the decision level of the integrated problem in the IHTC.
We observe that OT and nurses are integrated as primary resources seven times~\cite{rachuba2024integrated}. 
While the analysis in~\cite{rachuba2024integrated} did not explicitly include the patient-to-room assignment, we \edit{consider} beds and inpatient ward decisions \edit{as} closely related.
However, no existing papers combine these \edit{decisions} with both OT and nursing staff. 
Furthermore, no studies were identified that incorporate patient admissions as an alternative to patient-to-room allocation in this context.
Logically, a comprehensive model integrating operational decisions for those four resources is absent from the literature up to 2023.

This search yields 15 peer-reviewed journal papers.
Of these, five address integrated healthcare problems~\cite{bigharaz_integrated_2025, zanazzo_multi-neighborhood_2025, bigharaz_rolling_0000, brandt_integrated_2025, bikker_generic_2025} and are included in our analysis.
\edit{The remaining ten are excluded as they focus on literature reviews, qualitative surveys, strategic planning, or single-resource planning problems, which all fall outside the scope.}

The approaches in~\cite{bigharaz_integrated_2025,bigharaz_rolling_0000,bikker_generic_2025} focus on blueprint scheduling problems for appointment planning. They take the allocation of rooms and staff capacity into account. However, compared to the IHTC, the blueprint scheduling takes place \edit{at} a tactical level, \edit{aggregating} patient groups or types instead of individual patients.
The approaches in~\cite{brandt_integrated_2025} and~\cite{zanazzo_multi-neighborhood_2025} address integrated healthcare problems that consider similar resources as in the IHTC and they focus on an operational planning level.
In~\cite{brandt_integrated_2025}, the integrated patient-room and nurse-patient assignment is solved using an MIP (for small instances) and a greedy constructive heuristic that iterates through a list of patients each day to determine the patient-to-room and nurse-to-patient assignment simultaneously.
The assignment decision that yields the best contribution to the objective is chosen.
The same problem is addressed in~\cite{zanazzo_multi-neighborhood_2025} using \edit{a}n SA approach with four neighbourhood operators (room swap or change, nurse swap or change).

Within the literature reviewed in~\cite{rachuba2024integrated} and the additionally identified publications, MIP is the most widely used methodology, e.g., in~\cite{bigharaz_integrated_2025, brandt_integrated_2025, bikker_generic_2025, brandt2025patient}.
The authors of~\cite{brandt2025patient} demonstrate that a well-structured integer programming formulation can yield strong computational performance in patient admission and room assignment problems.
Their computational study further highlights the importance of model design in enhancing solver efficiency.
In line with~\cite{brandt2025patient}, a rigorous MIP formulation for the static operational patient admission scheduling problem was developed in~\cite{bastos2019mixed}, which demonstrates its effectiveness by generating new best-known solutions for benchmark instances and proves optimality for selected cases.
Further applied techniques are heuristics~\cite{rachuba2024integrated, bigharaz_rolling_0000, brandt_integrated_2025, zanazzo_multi-neighborhood_2025} and discrete event simulation~\cite{rachuba2024integrated, mokashi_simulation-based_2025}.
For example, the approach~\cite{zanazzo_multi-neighborhood_2025} shows the successful application of SA \edit{to} the integrated patient-room and nurse-patient assignment.
\edit{Although CP is not explicitly mentioned in~\cite{rachuba2024integrated}, it has been successful in hospital applications where the focus is to quickly identify a feasible solution, rather than finding optimal solutions. For example,~\cite{wang2015scheduling} uses CP for OT scheduling and~\cite{rios2025surgery} uses CP for surgery scheduling.}
%However, it has been shown to be successful in hospital applications when the focus is on finding feasible solutions fast compared to finding optimal solutions, see e.g., \cite{wang2015scheduling} for OT scheduling or~\cite{rios2025surgery} for surgery scheduling.

Based on the reviewed studies, we conclude that some studies integrate room and nurse assignment decisions, but they omit the OT and patient admission dimensions. Consequently, the specific combination of resources central to the IHTC remains unexplored in the literature until the start of the competition.
Furthermore, the combination of MIP, CP, and SA into one solution approach does not exist in the reviewed studies.

Combining different solution approaches in general, also outside of healthcare applications, has been shown to be successful in the past.
A very prominent example is logic-based Benders' decomposition~\cite{HookerO03}, which consists of a 2-stage approach with an MIP for solving the first stage and typically a CP for solving the second stage.

\section{Problem setting}
\label{sec_problemsetting}

\DeclareDocumentCommand\tikzDecisions{}{%
  \begin{tikzpicture}[baseline=-2pt]
    \node[rectangle, draw=blue, thick, fill=blue!20, align=center, minimum height=1em, minimum width=2.5em] {};
  \end{tikzpicture}%
}
\DeclareDocumentCommand\tikzSoftConstraints{}{%
  \begin{tikzpicture}[baseline=-2pt]
    \node[rectangle, dashed, draw=green!60!black, fill=green!40!white, thick, align=center, rounded corners, minimum height=1em, minimum width=2.5em] {};
  \end{tikzpicture}%
}
\DeclareDocumentCommand\tikzHardConstraints{}{%
  \begin{tikzpicture}[baseline=-2pt]
    \node[rectangle, draw=red, fill=red!20, thick, align=center, rounded corners, minimum height=1em, minimum width=2.5em] {};
  \end{tikzpicture}%
}

In this section, we give an overview of the planning problem including decisions, constraints and subproblems.
The details of the actual optimization problems will be introduced in the relevant sections in the remainder of the paper.
For a detailed description of the problem and the IHTC we refer to~\cite{IHTC24}.

\begin{figure}[htpb]
  \centering%
  \resizebox{0.8\textwidth}{!}{
  \begin{tikzpicture}[
    scale=0.98,
    decision/.style = {rectangle, draw=blue, thick, fill=blue!20, align=center},
    hard constraint/.style = {rectangle, draw=red, very thick, fill=red!20, align=center, rounded corners},
    soft constraint/.style = {rectangle, draw=green!60!black, very thick, fill=green!40!white, dashed, align=center, rounded corners},
    hard link/.style = {ultra thick, draw=red},
    soft link/.style = {ultra thick, draw=green!80!black},
    ]

  \node[decision, text width=5em] (admission) at (0,0) {Patient admission (day)};
  \node[decision, text width=5em] (rooms) at (0,-2.5) {Patient-to-room assignment};
  \node[decision, text width=5em] (theaters) at (4,0) {Patient-to-OT assignment};
  \node[decision, text width=6em] (nurses) at (4,-2.5) {Nurse-to-room assignment};
  \node[circle, fill=red!20!white, draw=red, inner sep=0mm, minimum size=2mm] (admission-rooms-nurses) at (2,-3) {};
  \node[circle, fill=red!20!white, draw=red, inner sep=0mm, minimum size=2mm] (admission-rooms) at (-0.5,-1.25) {};
  \node[circle, fill=red!20!white, draw=red, inner sep=0mm, minimum size=2mm] (admission-theaters) at (2,0) {};

  \draw[hard link] (admission) to (admission-rooms);
  \draw[hard link] (rooms) to (admission-rooms);
  \draw[hard link] (admission) to (admission-theaters);
  \draw[hard link] (theaters) to (admission-theaters);
  \draw[hard link] (admission) -- (admission-rooms-nurses);
  \draw[hard link] (rooms) -- (admission-rooms-nurses);
  \draw[hard link] (nurses) -- (admission-rooms-nurses);

  \node[hard constraint] (room-capacity) at (-3.8,-2.3) {Room capacity};
  \draw[hard link, ->] (admission-rooms) to (room-capacity.north east);

  \node[hard constraint, text width=23mm] (gender) at (-3.5,-3.5) {One gender per room};
  \draw[hard link, ->] (admission-rooms) to (gender);

  \node[hard constraint] (mandatory) at (-3.75,1.75) {Mandatory?};
  \draw[hard link, ->] (admission) to (mandatory.east);
  \node[hard constraint] (deadline) at (-3.5,1) {Deadline};
  \draw[hard link, ->] (admission) to (deadline.east);
  \node[hard constraint] (release) at (-3.5,0.25) {Release day};
  \draw[hard link, ->] (admission) to (release.east);
  \node[hard constraint] (surgeon-capacity) at (-3.5,-0.5) {Surgeon capacity};
  \draw[hard link, ->] (admission) to (surgeon-capacity.east);

  \node[hard constraint] (theater-capacity) at (3.4,1.85) {OT capacity};
  \draw[hard link, ->] (admission-theaters) to (theater-capacity.south);
  
  \node[soft constraint, text width=5em] (postponed) at (-1,1.85) {Postponed patient};
  \draw[soft link, ->] (admission) to (postponed);

  \node[soft constraint,text width=4em] (delayed) at (1,1.85) {Delayed patient};
  \draw[soft link, ->] (admission) to (delayed);

  \node[soft constraint, text width=5em] (surgeon-transfer) at (6.25,1.5) {Surgeon switching OTs};
  \draw[soft link, ->] (admission-theaters) to[bend left=15] (surgeon-transfer.west);

  \node[soft constraint, text width=4em] (opened-theaters) at (6.4,0) {Opened OTs};
  \draw[soft link, ->] (theaters) to (opened-theaters.west);

  \node[soft constraint, text width=7em] (age-room-mix) at (-3.9,-1.40) {Mix of age-groups in room};
  \draw[soft link, ->] (admission-rooms) to (age-room-mix.east);

  \node[soft constraint, text width=5em] (nurse-workload) at (-0.5,-4.5) {Excess workload};
  \draw[soft link, ->] (admission-rooms-nurses) to[bend right=8] (nurse-workload);

  \node[soft constraint, text width=5em] (nurse-skill) at (1.75,-4.5) {Skill level mismatch};
  \draw[soft link, ->] (admission-rooms-nurses) to (nurse-skill);

  \node[soft constraint, text width=5em] (continuity-of-care) at (4,-4.5) {Continuity of care};
  \draw[soft link, ->] (admission-rooms-nurses) to[bend left=8] (continuity-of-care);

  % \node[decision, opacity=0.5, minimum height=1.5em] at (-5.5, -5.0) {decisions};
  % \node[hard constraint, opacity=0.5, minimum height=1.5em] at (-3.4, -5.0) {hard constraint};
  % \node[soft constraint, opacity=0.5, minimum height=1.5em] at (0.1, -5.0) {soft constraint (with penalty)};
  \end{tikzpicture}}
  \caption{%
    Interplay of decisions (\tikzDecisions), soft- (\tikzSoftConstraints) and hard- (\tikzHardConstraints) constraints (OT = operating theater).
  }
  \label{fig_problem}
\end{figure}

The overall aim \edit{of the problem} is to admit \edit{as many} elective patients \edit{as possible} to a hospital, while taking into account the available capacity of OTs, rooms, and nurses.
A solution to the planning problems addresses four decisions over a planning horizon of several weeks.
The first decisions are the admission days of the patients.
Additionally, patients need to be assigned to an OT on the admission day and a patient room for the (known) length of stay.
Finally, given a known staff-to-shift roster, the nurses need to be assigned to patient rooms during their shifts to care for patients.
Some of the patients are mandatory and have to be admitted during the considered planning horizon, \edit{whereas} others can be postponed to later planning horizons.
Furthermore, already admitted patients from previous periods (called occupants) have to be taken into account.

The decisions are restricted by hard constraints such as time windows for patient admissions, available capacities of surgeons, rooms, and operating theaters, as well as the predetermined nurse schedule.
Furthermore, patients of different genders cannot share a room.

The quality of a solution depends on a range of soft constraints that consider patient- and staff-related factors.\edit{
 These soft constraints are enforced using penalty terms in the objective.
The first soft constraint penalizes having different patient age groups in one room.
A continuity-of-care soft constraint induces a penalty term for each different nurse that treats a patient.
A penalty is also occurred for each delayed or postponed patient, and for each time period where a patient is treated by a nurse with an insufficient skill level.
Furthermore, there is a penalty for each nurse and time period where the nurse has excess workload.
Additionally, opening an OT and switching surgeons between different OTs incurs additional costs.
These eight objectives are balanced using different weights in the objective, one for each penalty category.
}%
\edit{In the competition, the} weight of each soft constraint violation varies across instances, making an adaptable solution approach necessary.
The interplay between decisions and constraints is depicted in \cref{fig_problem}.

\DeclareDocumentCommand\days{}{\ensuremath{\mathcal{D}}}
\DeclareDocumentCommand\shifts{}{\ensuremath{\mathcal{S}}}
\DeclareDocumentCommand\patients{}{\ensuremath{\mathcal{P}}}
\DeclareDocumentCommand\mandatoryPatients{}{\ensuremath{\mathcal{P}^{\textup{mand}}}}
\DeclareDocumentCommand\optionalPatients{}{\ensuremath{\mathcal{P}^{\textup{opt}}}}
\DeclareDocumentCommand\occupantPatients{}{\ensuremath{\mathcal{P}^{\textup{occ}}}}
\DeclareDocumentCommand\flexiblePatients{}{\ensuremath{\mathcal{P}^{\textup{flex}}}}
\DeclareDocumentCommand\clients{}{\ensuremath{\mathcal{C}}}
\DeclareDocumentCommand\nurses{}{\ensuremath{\mathcal{N}}}
\DeclareDocumentCommand\surgeons{}{\ensuremath{\mathcal{U}}}
\DeclareDocumentCommand\rooms{}{\ensuremath{\mathcal{R}}}
\DeclareDocumentCommand\genders{}{\ensuremath{\mathcal{G}}}
\DeclareDocumentCommand\ageGroups{}{\ensuremath{\mathcal{A}}}
\DeclareDocumentCommand\theaters{}{\ensuremath{\mathcal{O}}}

\DeclareDocumentCommand\releaseDay{m}{\ensuremath{d^{\textup{release}}_{#1}}}
\DeclareDocumentCommand\admissionDays{m}{\ensuremath{\mathcal{D}^{\textup{adm}}_{#1}}}
\DeclareDocumentCommand\roomCapacity{m}{\ensuremath{K^{\textup{room}}_{#1}}}
\DeclareDocumentCommand\theaterCapacity{mm}{\ensuremath{K^{\textup{thea}}_{#1,#2}}}
\DeclareDocumentCommand\surgeonCapacity{mm}{\ensuremath{K^{\textup{surg}}_{#1,#2}}}
\DeclareDocumentCommand\nurseCapacity{mmm}{\ensuremath{K^{\textup{nurse}}_{#1,#2,#3}}}
\DeclareDocumentCommand\lengthStay{m}{\ensuremath{\ell^{\textup{stay}}_{#1}}}
\DeclareDocumentCommand\durationSurgery{m}{\ensuremath{T^{\textup{surg}}_{#1}}}
\DeclareDocumentCommand\durationPatient{mmm}{\ensuremath{T^{\textup{pat}}_{#1,#2,#3}}}
\DeclareDocumentCommand\skillPatient{mmm}{\ensuremath{\sigma^{\textup{pat}}_{#1,#2,#3}}}
\DeclareDocumentCommand\skillNurse{m}{\ensuremath{\sigma^{\textup{nurse}}_{#1}}}
\DeclareDocumentCommand\genderPatient{m}{\ensuremath{\gamma^{\textup{pat}}_{#1}}}
\DeclareDocumentCommand\ageGroupPatient{m}{\ensuremath{\alpha^{\textup{pat}}_{#1}}}
\DeclareDocumentCommand\careBound{mm}{\ensuremath{\beta^{\textup{care}}_{#1,#2}}}

\DeclareDocumentCommand\presentPatients{}{\ensuremath{\mathcal{P}^{\textup{present}}}}
\DeclareDocumentCommand\presentDays{m}{\ensuremath{\mathcal{D}_{#1}^{\textup{present}}}}
\DeclareDocumentCommand\nursesShift{mm}{\ensuremath{\mathcal{N}_{#1,#2}^{\textup{present}}}}

\begin{table}[htpb]
  \caption{Nomenclature for the instance data: sets.}
  \label{tab_nomenclature_sets}
  \centering%
  \footnotesize
  {\renewcommand{\arraystretch}{1.1}%
  \begin{tabular}{r@{\;}l|p{0.62\textwidth}|r}\toprule
    \multicolumn{2}{c|}{\textbf{Symbol}} & \textbf{Definition} & \textbf{Subproblems} \\ \midrule
    \days & $\subset \Znonneg$ & Days in planning horizon $\{1,2,\dotsc\}$ & all \\
    \patients & & Patients &  all \\
    \mandatoryPatients & $\subseteq \patients$ & Mandatory patients & all \\
    \optionalPatients & $\subseteq \patients$ & Optional patients & all \\
    \occupantPatients & $\subseteq \patients$ & Occupants, i.e. already admitted patients & all \\
    \flexiblePatients & $\subseteq \patients$ & Patients to admit $\flexiblePatients \coloneqq \mandatoryPatients \cup \optionalPatients$ & all \\
    \nurses & & Nurses & all \\
    \shifts & & Shifts: $\shifts = \{ \textup{early}, \textup{late}, \textup{night} \}$ & all \\
    \surgeons & & Surgeons & all \\
    \rooms & & Rooms & all \\
    \genders & & Genders: $\genders = \{ \textup{female}, \textup{male} \}$ & all \\
    \ageGroups & & Age groups & all \\
    \theaters & & Operating theaters & all \\
\midrule
    \presentPatients & $\subseteq \patients$ & Admitted patients (including occupants) & 2.--4. \\
    $\presentDays{p}$ & $\subseteq \days$ & Days on which patient $p \in \presentPatients$ is present & 2.--4. \\
    $\nursesShift{d}{s}$ & $\subseteq \nurses$ & Nurses available on day $d \in \days$ during shift $s \in \shifts$ & all\\
\bottomrule
  \end{tabular}}
\end{table}

\begin{table}[htpb]
  \caption{Nomenclature for the instance data: parameters.}
  \label{tab_nomenclature_parameters}
  \centering%
  \footnotesize
  {\renewcommand{\arraystretch}{1.05}%
  \begin{tabular}{r@{\;}l|p{0.74\textwidth}}\toprule
    \multicolumn{2}{c|}{\textbf{Symbol}} & \textbf{Definition}  \\ \midrule
    $\releaseDay{p}$ & $\in \days$ & Release day of patient $p \in \flexiblePatients$ \\
    $\admissionDays{p}$ & $\subseteq \days$ & Possible admission days of patient $p \in \flexiblePatients$ \\
    $\roomCapacity{r}$ & $\in \Znonneg$ & Room capacity of room $r \in \rooms$ \\
    $\theaterCapacity{o}{d}$ & [mins] & Capacity of operating theater $o \in \theaters$ on day $d \in \days$. \\
    $\surgeonCapacity{u}{d}$ & [mins] & Capacity of surgeon $u \in \surgeons$ on day $d \in \days$. \\
    $\nurseCapacity{n}{d}{s}$ & [mins] & Capacity of nurse $n \in \nurses$ on day $d \in \days$ during shift $s \in \shifts$. \\
    $\lengthStay{p}$ & $\in \Znonneg$ & Length of stay of patient $p \in \flexiblePatients$ in days \\
    $\durationSurgery{p}$ & [mins] & Duration of surgery of patient $p \in \flexiblePatients$ \\
    $\durationPatient{p}{\delta}{s}$ & [mins] & Workload incurred by patient $p \in \patients$ on day $(d + \delta)$ during shift $s \in \shifts$ if admitted on $d \in \days$; 0 if $\delta \notin \{0,1,\dotsc,\lengthStay{p}-1\}$ \\
    $\skillPatient{p}{\delta}{s}$ & $\in \Z$ & Skill level required by patient $p \in \patients$ on day $(d + \delta)$ during shift $s \in \shifts$ if admitted on $d \in \days$; $\delta \in \{0,1,\dotsc,\lengthStay{p}-1\}$ \\
    $\skillNurse{n}$ & $\in \Z$ & Skill level of nurse $n \in \nurses$ \\
    $\genderPatient{p}$ & $\in \genders$ & Gender of patient $p \in \patients$ \\
    $\ageGroupPatient{p}$ & $\in \ageGroups$ & Age group of patient $p \in \patients$ \\
    $\careBound{p}{d}$ & $\in \Znonneg$ & Lower bound on care costs if patient $p \in \flexiblePatients$ is admitted on day $d \in \days$ \\ \bottomrule
  \end{tabular}}
\end{table}

\DeclareDocumentCommand\weightUnsched{}{\ensuremath{\omega^{\textup{unsched}}}}
\DeclareDocumentCommand\weightDelay{}{\ensuremath{\omega^{\textup{delay}}}}
\DeclareDocumentCommand\weightTheater{}{\ensuremath{\omega^{\textup{thea}}}}
\DeclareDocumentCommand\weightSurgeon{}{\ensuremath{\omega^{\textup{surg}}}}
\DeclareDocumentCommand\weightAgemix{}{\ensuremath{\omega^{\textup{agmx}}}}
\DeclareDocumentCommand\weightWorkload{}{\ensuremath{\omega^{\textup{wkld}}}}
\DeclareDocumentCommand\weightCoC{}{\ensuremath{\omega^{\textup{coc}}}}
\DeclareDocumentCommand\weightSkill{}{\ensuremath{\omega^{\textup{skill}}}}

\begin{table}[htpb]
  \caption{Nomenclature for the instance data: objective weights.}
  \label{tab_nomenclature_weights}
  \footnotesize
  \centering%
  {\renewcommand{\arraystretch}{1.05}%
  \begin{tabular}{r@{\;}l|p{0.74\textwidth}}\toprule
    \multicolumn{2}{c|}{\textbf{Symbol}} & \textbf{Meaning}  \\ \midrule
    \weightUnsched & $\in \Znonneg$ & Objective cost for each unscheduled patient \\
    \weightDelay & $\in \Znonneg$ & Objective cost for each day admission is delayed (if admitted) \\
    \weightTheater & $\in \Znonneg$ & Objective cost for opening theaters \\
    \weightSurgeon & $\in \Znonneg$ & Objective cost per surgeon transfer \\
    \weightAgemix & $\in \Znonneg$ & Objective cost for age-mix per room and day \\
    \weightWorkload & $\in \Znonneg$ & Objective cost for nurses' excess workload \\
    \weightCoC & $\in \Znonneg$ & Objective cost for total continuity of care (number of nurses taking care of each admitted patient) \\
    \weightSkill & $\in \Znonneg$ & Objective cost for skill level mismatch per shift and patient\\\bottomrule
  \end{tabular}}
\end{table}

\DeclareDocumentCommand\surgeonOf{m}{\ensuremath{u(#1)}}
\DeclareDocumentCommand\dayOf{m}{\ensuremath{d(#1)}}
\DeclareDocumentCommand\roomOf{m}{\ensuremath{r(#1)}}

\begin{table}[htpb]
  \caption{Nomenclature: assignments.}
  \label{tab_nomenclature_assignments}
  \centering%
  \footnotesize  {\renewcommand{\arraystretch}{1.05}%
  \begin{tabular}{r@{\;}l|p{0.64\textwidth}|r}\toprule
    \multicolumn{2}{c|}{\textbf{Symbol}} & \textbf{Meaning} & \textbf{Subproblem} \\ \midrule
    \surgeonOf{p} & $\in \surgeons$ & The surgeon of patient $p \in \flexiblePatients$ & all \\
    \dayOf{p} & $\in \days$ & The admission day of patient $p \in \presentPatients$; $d(p) = 1$ for all $p \in \occupantPatients$ & 2.--4. \\
    \roomOf{p} & $\in \rooms$ & The room assigned to patient $p \in \presentPatients$ & 3.--4. \\\bottomrule
  \end{tabular}}
\end{table}

\section{Solution approach}
\label{sec_approach}

The integrated planning problem \edit{involves making} all four decisions presented in \cref{sec_problemsetting} \edit{
so as to} optimize the overall objective function.
Initial experiments showed that solving larger instances using one monolithic MIP is impossible within the given time limit of 10~minutes.
Therefore, we decompose the planning problem into the following smaller subproblems:

\begin{enumerate}
\item
  \emph{Patient admission}: decides on which day a patient should be admitted or if their admission should be postponed to the next decision horizon.
\item
  \emph{Patient-to-room assignment}: allocates each admitted patient (from 1.) to a room for the entire length of stay.
\item
  \emph{Patient-to-OT assignment}: allocates each admitted patient (from 1.) to an OT on the day of admission.
\item
  \emph{Nurse-to-room assignment}: assigns rooms to available nurses during each shift, taking the patient-to-room assignment from 2. as input.
\end{enumerate}

The choice for this specific decomposition is based on the problem setting.
As can be seen in \cref{fig_problem}, if the admission day of the patients is known, the patient-to-OT assignment is independent of the patient-to-room and nurse-to-room assignments.
Furthermore, once the patient-to-room assignment is known, all nurse-to-room assignments are feasible, since they only involve soft constraints.

The nomenclature used in this paper is defined in \cref{tab_nomenclature_sets,tab_nomenclature_parameters,tab_nomenclature_weights,tab_nomenclature_assignments} where the column \textit{Subproblem} relates to the relevant subproblems of our solution approach as indicated by the number in the enumeration above.

Our approach uses a combination of MIP, CP, and SA to solve the overall problem and consists of three phases.
An overview of the overall solution approach is given in \cref{fig:solution_approach}, including the three phases and their exchange of information.
The main purpose of Phase~1 is to compute, for every possible assignment of a single patient to a specific admission day, a lower bound on the nurse-to-room assignment costs that this would incur.
In parallel, our approach already begins to solve the MIP for patient admission (with trivial lower bounds).
The phase ends as soon as these lower bounds are computed. 
\edit{Consequently,} all MIP solutions that were found so far are used to warm-start the admission MIP in Phase~2.
Phase~2 is the core phase, completing solutions to the overall integrated planning problem by iteratively solving patient admission, patient-to-room assignment, patient-to-OT assignment, and nurse-to-room assignment.
Phase~3 aims at improving the nurse-to-room assignment using an exact approach.
In the remainder of this section, we explain the phases, including their mathematical formulations, solution methods, and design choices in more detail.
Furthermore, \cref{sec_rejection} presents a variant of the patient admission problem that we developed after the competition to generate more feasible admission schedules.

\begin{figure}[htpb]
  \resizebox{\textwidth}{!}{
  \begin{tikzpicture}[
    xshift=35mm,
    yshift=-1mm,
    mip/.style = {rectangle, draw=red, thick, fill=red!20, align=center},
    cp/.style = {rectangle, draw=blue, thick, fill=blue!20, align=center, dashed},
    sa/.style = {rectangle, draw=green!60!black, thick, fill=green!20, align=center, rounded corners},
    ]

    \node at (0,-0.1) {Thread \#1};
    \node at (3,-0.1) {Thread \#2};
    \node at (6,-0.1) {Thread \#3};
    \node at (9,-0.1) {Thread \#4};

    \node[mip, text width=26mm, minimum height=15mm] (admi) at (0,-0.5) {Initial \\ patient \\ admission MIP};
    \node[mip, text width=26mm, minimum height=15mm] (care1) at (3,-0.5) {Care-cost \\ lower bound \\MIP};
    \node[mip, text width=26mm, minimum height=15mm] (care2) at (6,-0.5) {Care-cost \\ lower bound \\MIP};
    \node[mip, text width=26mm, minimum height=15mm] (care3) at (9,-0.5) {Care-cost \\ lower bound \\MIP};

    \draw[very thick,dotted] (-3.8,-2.3) -- (11.7,-2.3);
    \draw[very thick,dotted] (-3.8,-7.1) -- (11.7,-7.1);
    \node[rotate=90] at (-3.6,-1) {Phase 1};
    \node[rotate=90] at (-3.6,-4.0) {Phase 2};
    \node[rotate=90] at (-3.6,-8.0) {Phase 3};

    \node[mip, text width=26mm, minimum height=8mm] (adm1) at (0,-2.9) {Patient\\ admission MIP};
    \node[mip, text width=26mm, minimum height=18mm] (adm2) at (0,-4.8) {Patient\\ admission MIP \\ with reduced \\ aggregated \\ room capacity};

    \node[cp, text width=50mm, minimum height=28mm] (room) at (4.5,-3.9) {Patient-to-room assignment CP \\[8mm] If infeasible several times, \\ reduce aggregated room \\ capacity in admission MIP.};
    \node[mip, text width=50mm, minimum height=8mm] (theater) at (4.5,-6.4) {Patient-to-OT assignment MIPs};

    \node[sa, text width=26mm, minimum height=28mm] (nurs) at (9,-3.9) {Nurse-to-room SA};

    \node[mip, text width=26mm, minimum height=8mm] (final1) at (0,-8.4) {Nurse-to-room MIP};
    \node[mip, text width=26mm, minimum height=8mm] (final2) at (3,-8.4) {Nurse-to-room MIP};
    \node[mip, text width=26mm, minimum height=8mm] (final3) at (6,-8.4) {Nurse-to-room MIP};
    \node[mip, text width=26mm, minimum height=8mm] (final4) at (9,-8.4) {Nurse-to-room MIP};

    \draw[ultra thick,->,orange] (admi) -- +(-2.5,0) |- node[pos=0.27,left] (admisol) {sol} (adm1);
    \draw[ultra thick,->,purple] (care1) |- node[pos=0.75,above] {cost update} (admisol);
    \draw[ultra thick,->,purple] (care2) |- (admisol);
    \draw[ultra thick,->,purple] (care3) |- (admisol);

    \draw[ultra thick,->,orange] (adm1) -- +(1.9,0);
    \draw[ultra thick,->,orange] (adm2) -- +(1.9,0);
    % \draw[ultra thick,->,orange] (room) -- +(3.1,0);
    \draw[ultra thick,->,orange] (nurs) -- +(1.9,0) node[black, anchor=west, align=center] (solpool) {sol \\ pool};
    \draw[ultra thick,->,orange] (room) -- +(0,-2.1);
    \draw[ultra thick,->,orange] (theater) -| +(4.5,1.1);

    \draw[ultra thick,orange] (solpool) -- +(0,-3.7) node[inner sep=0,draw=orange,fill=orange, minimum size=0.01mm] (bestsols) {};
    \draw[ultra thick,orange,->] (bestsols) -| node[pos=0.29,above] {best 4 solutions} (final1);
    \draw[ultra thick,orange,->] (bestsols) -| (final4);
    \draw[ultra thick,orange,->] (bestsols) -| (final2);
    \draw[ultra thick,orange,->] (bestsols) -| (final3);

  \end{tikzpicture}}
  \caption{%
    Overview of the phases of the solution approaches including allocation to threads and information exchange.
    Solid boxes indicate MIPs, the dashed box indicates a constraint-programming approach and the rounded corners indicate the SA approach.
    Arrows indicate the flow of (partial) solution information.
  }
  \label{fig:solution_approach}
\end{figure}

\subsection{Preprocessing}
\label{sec_approach_preprocessing}

Before starting Phase~1, we preprocess the instance to exclude obviously infeasible or suboptimal solutions.
In fact, we compute the set $\admissionDays{p}$ of potential admission days for each patient $p \in \flexiblePatients$ by considering the release day, the deadline and by removing days $d$ on which the patient's surgeon has no time, that is, for which $\surgeonCapacity{\surgeonOf{p}}{d} < \durationSurgery{p}$ holds.
Finally, we remove candidate admission days for which the total cost for delay and continuity of care exceeds the cost for postponing admission.

\subsection{Phase~1: Lower bounds for patient admission}
\label{sec_approach_carebounds}

Due to the decomposition into the four subproblems as stated in Section \ref{sec_approach}, the patient admission problem only includes room, surgeon, and OT capacity on an aggregated level and ignores more detailed information on later subproblems, e.g., the cost of nurse-to-room assignments.
Without this information, promising admission days for patients are hard to identify.
To overcome this, we calculate lower bounds for combinations of patients and admission days by using information from later subproblems.

Thus, during Phase~1, we only use one thread for solving the patient admission MIP \eqref{model_admission_obj}--\eqref{eq:adm_room} presented in \cref{sec_approach_decomposition_admission} to start finding patient admission solutions.
The other three threads solve another MIP  to calculate lower bounds $\careBound{p^\star}{d^\star}$ on the cost incurred when patient $p^\star \in \flexiblePatients$ is admitted on day $d^\star\in \admissionDays{p^\star}$.
\edit{The purpose of these lower bounds is} to find more promising patient admission solutions utilizing the information from later subproblems.

We specifically look at the nurse-to-room assignment that is governed by soft constraints on skill level, excess workload and continuity of care.
It is very often possible to find a nurse with the required skill level for each day and shift during which a patient $p^\star$ is admitted, leading to low skill-mismatch costs.
Moreover, it is also often possible to cover the whole period of stay of a patient by only three nurses, which would result in lower continuity of care costs.
However, this will lead to a very high excess workload for nurses.
This means that there is a trade-off between these \edit{two} soft constraints \edit{in the sense that a minimum-penalty solution for one of them is structurally very different from a solution that minimizes the other penalty.}
This motivates Phase~1.

To calculate $\careBound{p^\star}{d^\star}$, we solve the MIP \eqref{eq:bound_obj}--\eqref{eq:bound_coc} for each $p^\star \in \flexiblePatients$ and $d^\star \in \admissionDays{p}$.
For each pair $(p^\star,d^\star)$, the availability of nurses and the level of skills on the corresponding days on which $p^\star$ would stay incur a minimum sum of skill mismatch and continuity of care penalties, which can be used as a lower bound.

\DeclareDocumentCommand\varNurseDayShift{ooo}{\ensuremath{\IfValueTF{#3}{x_{#1,#2,#3}}{x}}}
\DeclareDocumentCommand\varNurse{o}{\ensuremath{\IfValueTF{#1}{y_{#1}}{y}}}

We use the decision variables $\varNurseDayShift[n][d][s] \in \{0,1\}$ to indicate whether nurse $n \in \nurses$ cares for $p^\star$ on day $d \in \days$ and during shift $s \in \shifts$ ($\varNurseDayShift[n][d][s] = 1$) or not ($\varNurseDayShift[n][d][s] = 0$).
Note that if nurse $n$ does not work during shift $s$ on day $d$, $\varNurseDayShift[n][d][s]$ is a parameter equal to $0$.
Moreover, the variables $\varNurse[n] \in \{0,1\}$ state whether nurse $n \in \nurses$ cares for patient $p^\star$ at least once ($\varNurse[n] = 1$) or never ($\varNurse[n] = 0$).

The objective~\eqref{eq:bound_obj} is to minimize the weighted sum of skill \edit{mismatch penalties} and continuity of care costs.
Here, the set \presentDays{p^\star} contains all days that the patient will be present when admitted on day $d^\star$.
\begin{equation}
  \weightSkill \sum_{n \in \nurses} \sum_{d \in \presentDays{p^\star}} \sum_{s \in \shifts} \varNurseDayShift[n][d][s] + \weightCoC \sum_{n \in \nurses} \varNurse[n] \label{eq:bound_obj}
\end{equation}

Inequalities~\eqref{eq:bound_minnurse} enforce that (at least) one nurse takes care of the patient in every shift during the patient's stay, and inequalities~\eqref{eq:bound_coc} make sure that variables $\varNurse[n]$ track if nurse $n$ is assigned to the patient in any shift.
\begin{equation}
  \sum_{n \in \nurses} \varNurseDayShift[n][d][s] \geq 1 \quad \forall d \in \presentDays{p^\star} ~ \forall s \in \shifts \label{eq:bound_minnurse}
\end{equation}
\begin{equation}
  \varNurse[n] \geq \varNurseDayShift[n][d][s] \quad \forall n \in \nurses ~ \forall d \in \presentDays{p^\star} ~ \forall s \in \shifts \label{eq:bound_coc}
\end{equation}
The optimum of this MIP defines the value of $\careBound{p^\star}{d^\star}$, which is then used in the objective function~\eqref{model_admission_obj} of the admission MIP in Phase~2 (see \cref{sec_approach_decomposition_admission}).
The patient admission MIP running on one thread during Phase~1 uses the trivial lower bound $\careBound{p}{d} = 3$ on continuity of care (since each patient requires at least three nurses due to three shifts per day).

Phase~1 ends as soon as all $\careBound{p^\star}{d^\star}$ parameters have been computed.
Then the lower bounds on costs of all feasible admission solutions that were found by the first thread in the meantime will be updated accordingly, and the solution approach continues with Phase~2.

\subsection{Phase~2: Decomposition}
\label{sec_approach_decomposition}

Phase~2 is the main phase and finds solutions to the overall integrated problem by iteratively solving patient admission, patient-to-room assignment, patient-to-OT assignment, and nurse-to-room assignment.
Phase~2 runs in parallel utilizing all four available threads\footnote{The IHTC 2024 competition rules allowed four threads.}, maintaining a pool of partial solutions.
Whenever there is a feasible solution to a subproblem, it is used as input to the next subproblem to be further completed.
The selection of which solution to proceed next for a subproblem is based on its objective value, considering all components of the objective function that can be determined already.
In the remainder of this section, we present the solution method for each subproblem.

\subsubsection{Patient admission}
\label{sec_approach_decomposition_admission}

First, we compute promising solutions to the patient admission problem.
For this, one thread is used for solving the MIP formulation \eqref{model_admission_obj}--\eqref{eq:adm_room} described below. 

\DeclareDocumentCommand\varAdmit{oo}{\ensuremath{\IfValueTF{#1}{x_{#1,#2}}{x}}}
\DeclareDocumentCommand\varStay{oo}{\ensuremath{\IfValueTF{#1}{z_{#1,#2}}{z}}}
\DeclareDocumentCommand\varPostpone{o}{\ensuremath{\IfValueTF{#1}{\pi_{#1}}{\pi}}}
\DeclareDocumentCommand\varExcess{oo}{\ensuremath{\IfValueTF{#1}{\varepsilon_{#1,#2}}{\varepsilon}}}
\DeclareDocumentCommand\varOpened{oo}{\ensuremath{\IfValueTF{#2}{\theta_{#1,#2}}{\IfValueTF{#1}{\theta_{#1}}{\theta}}}}

The patient admission MIP uses the decisions variables $\varAdmit[p][d] \in \{0,1\}$ for each patient~$p \in \flexiblePatients$ and  day~$d \in \admissionDays{p}$ to indicate whether $p$ is admitted on day $d$ ($\varAdmit[p][d] = 1$) or not ($\varAdmit[p][d] = 0$).
Note that we also use $\varAdmit[p][d]$ for $d \notin \admissionDays{p}$, but then it is a parameter equal to $0$.
The auxiliary variables $\varStay[p][d] \in \{0,1\}$ for each $p \in \patients$ and $d \in \days$ state whether patient $p$ is present on day $d$ ($\varStay[p][d] = 1$) or not ($\varStay[p][d] = 0$) and variables $\varPostpone[p] \in \{0,1\}$ indicate whether patient $p \in \optionalPatients$ is postponed ($\varPostpone[p] = 1$) or admitted ($\varPostpone[p] = 0$).
The variables $\varExcess[d][s] \in \Rnonneg$ represent (a lower bound on) the excess workload during shift $s \in \shifts$ on day $d \in \days$.
Finally, variables $\varOpened[o][d] \in \{0,1\}$ state whether operating theater $o \in \theaters$ is opened on day $d \in \days$ ($\varOpened[o][d] = 1$) or not ($\varOpened[o][d] = 0$).

The objective~\eqref{model_admission_obj} minimizes the sum of costs for delaying or postponing patients, excess workload, and opened theaters.
Additionally, we include the estimated lower bound on cost \careBound{p}{d} from Phase~1 for admitting patient $p$ on day $d$ (see \cref{sec_approach_carebounds}).
\begin{multline}
  \min \sum_{p \in \flexiblePatients} \sum_{d \in \admissionDays{p}} \left( \careBound{p}{d} + \weightDelay (d - \releaseDay{p}) \right) \varAdmit[p][d] \\
  + \weightUnsched \sum_{p \in \optionalPatients} \varPostpone[p]
  + \weightWorkload \sum_{d \in \days} \sum_{s \in \shifts} \varExcess[d][s]
  + \weightTheater \sum_{o \in \theaters} \sum_{d \in \days} \varOpened[o][d].
  \label{model_admission_obj}
\end{multline}
Equations~\eqref{eq:adm_linkz} and~\eqref{eq:adm_linkz2} ensure that the $\varStay$-variables representing the presence of a patient are linked correctly to the admission variables  $\varAdmit$.
\begin{alignat}{7}
  \varStay[p][d] &= \sum_{\delta = \max\{0,~ d - \lengthStay{p}+1\}}^{d} \varAdmit[p][\delta] &\qquad& \forall p \in \flexiblePatients ~ \forall d \in \days\label{eq:adm_linkz} \\
  \varStay[p][d] &= \begin{cases} 1 & \text{ if $ d \leq \lengthStay{p}$} \\ 0 & \text{ otherwise} \end{cases}  &\qquad& \forall p \in \occupantPatients ~ \forall d \in \days \label{eq:adm_linkz2}
\end{alignat}
Equations~\eqref{eq:adm_optional} and~\eqref{eq:adm_mandatory} enforce that optional patients are either postponed or admitted exactly once and mandatory patients are admitted, respectively.
\begin{alignat}{7}
  \sum_{d \in \admissionDays{p}} \varAdmit[p][d] &\;+\;& \varPostpone[p] &= 1 &\qquad& \forall p \in \optionalPatients \label{eq:adm_optional}\\
  \sum_{d \in \admissionDays{p}} \varAdmit[p][d] && &= 1 &\qquad& \forall p \in \mandatoryPatients \label{eq:adm_mandatory}
\end{alignat}
Inequalities~\eqref{eq:adm_surgery} ensure that the total surgery workload for each surgeon on each day does not exceed the surgeon’s capacity.
Inequalities~\eqref{eq:adm_ot} \edit{determine the number of opened OTs} depending on the needed aggregated surgery time, while also limiting it to the total available OT capacity.
\begin{equation}
  \sum_{\substack{p \in \flexiblePatients \\ u(p) = u}} \durationSurgery{p} \varAdmit[p][d] \leq \surgeonCapacity{u}{d} \qquad \forall u \in \surgeons ~ \forall d \in \days \label{eq:adm_surgery}
\end{equation}
\begin{equation}
  \sum_{o \in \theaters} \theaterCapacity{o}{d} \varOpened[o][d] \geq \sum_{p \in \flexiblePatients} \durationSurgery{p} \varAdmit[p][d] \quad \forall d \in \days \label{eq:adm_ot}
\end{equation}
Inequalities~\eqref{eq:adm_excess1} and~\eqref{eq:adm_excess2} make sure that the excess nurse workload aggregated over all nurses is a (non-negative) lower bound on the actual demand on each day and shift.
\begin{alignat}{7}
  \varExcess[d][s] &\geq 0 &\quad& \forall d \in \days ~ \forall s \in \shifts \label{eq:adm_excess1} \\
  \varExcess[d][s] + \sum_{n \in \nurses} \nurseCapacity{n}{d}{s} &\geq \hspace{-1ex} \sum_{p \in \flexiblePatients} \sum_{\hat{d} \in \days} \durationPatient{p}{d - \hat{d}}{s} \varAdmit[p][\hat{d}] + \hspace{-1ex} \sum_{p \in \occupantPatients} \hspace{-1ex} \durationPatient{p}{d}{s} &\quad& \forall d \in \days ~ \forall s \in \shifts\label{eq:adm_excess2}
\end{alignat}
Inequalities~\eqref{eq:adm_room} limit the total number of occupied beds on each day to the total number of available beds reduced by the parameter $\varrho \in \Znonneg$, which is described below.
\begin{equation}
   \sum_{p \in \patients} \varStay[p][d] \leq \left( \sum_{r \in \rooms} \roomCapacity{r} \right) - \varrho \qquad \forall d \in \days \label{eq:adm_room}
\end{equation}
The patient-admission MIP is solved with Gurobi~\cite{GUROBI}. 
Whenever a new feasible solution is found by the solver (using the solver callback), it is added to the pool of admission solutions, where it will be picked up for patient-to-room assignment (see \cref{sec_approach_decomposition_room}).
Since constraint \eqref{eq:adm_room} only includes room capacity on an aggregated level, the patient admission solution could lead to infeasible patient-to-room assignments, in particular, due to the gender constraints.

To increase the chance of finding feasible patient-to-room assignments, we lower this aggregated room capacity by the parameter $\varrho \in \Znonneg$ (starting with $\varrho = 0$).
We increase $\varrho$ iteratively by $1$, if no feasible patient-to-room assignments are found for a given number of tries.
Each admission solution is tagged with its value for $\varrho$, where those with a larger $\varrho$-value have a higher chance of finding a feasible patient-room assignment (since the adjusted aggregated room capacity is lower than the actual aggregated room capacity).
If too many admission solutions with the same $\varrho$-value cannot be augmented with a feasible patient-to-room assignment, consideration of an admission solution with this $\varrho$-value is penalized.
More precisely, if for six admission solutions with $\varrho = k$ no feasible solution is found within the working limits, then the admission MIP is restarted with $\varrho = k+1$.
\edit{This capacity reduction may reduce the overall solution quality}, but makes it easier to find feasible solutions in subsequent subproblems.

Finally, note that we obtain a valid lower bound on the optimum from the dual bound of the MIP (only) as long as $\varrho = 0$ holds.

\subsubsection{Patient-to-room assignment}
\label{sec_approach_decomposition_room}

Feasible solutions from the patient admission problem are sent to the patient-to-room assignment.

\DeclareDocumentCommand\varPatientRoom{oo}{\ensuremath{\IfValueTF{#1}{x_{#1,#2}}{x}}}
\DeclareDocumentCommand\varRoomDayGender{ooo}{\ensuremath{\IfValueTF{#1}{y_{#1,#2,#3}}{y}}}
\DeclareDocumentCommand\varRoomDayAge{ooo}{\ensuremath{\IfValueTF{#1}{\alpha_{#1,#2,#3}}{\alpha}}}
\DeclareDocumentCommand\varRoomDayAgediff{oo}{\ensuremath{\IfValueTF{#1}{\Delta_{#1,#2}}{\mu}}}

In this subproblem, we already assigned an admission day to each admitted patient.
Therefore, we define the set $\presentPatients \subseteq \patients$ (with $\mandatoryPatients \cup \occupantPatients \subseteq \presentPatients$) as the combined set of admitted patients and occupants\footnote{Patients already present at the beginning of the planning horizon.}.
The admission day $d$ of a patient $p \in \presentPatients$ itself is not so relevant for the room assignment, but the days $\presentDays{p} \coloneqq \{d, d+1, \dotsc, d + \lengthStay{p} - 1\} \cap \days$ (with $d \coloneqq 1$ if $p \in \occupantPatients$) on which patient $p$ is present matters.

We use decision variables $\varPatientRoom[p][r] \in \{0,1\}$ to indicate whether patient $p \in \presentPatients$ is assigned to room $r \in \rooms$ ($\varPatientRoom[p][r] = 1$) or not ($\varPatientRoom[p][r] = 0$).
Note that if room $r$ is incompatible with patient $p$, this will just be a parameter equal to $0$.
Moreover, we use variables $\varRoomDayGender[r][d][g] \in \{0,1\}$ to indicate whether room $r$ is used for patients of gender $g$ on day $d$ ($\varRoomDayGender[r][d][g] = 1$) or not ($\varRoomDayGender[r][d][g] = 0$).
To model age differences, we first introduce binary variables $\varRoomDayAge[r][d][a] \in \{0,1\}$ for all rooms $r \in \rooms$, days $d \in \days$ and age groups $a \in \ageGroups$ to indicate whether room $r$ has patients of age group $a$ on day $d$ or not ($\varRoomDayAge[r][d][a] = 0$).
Second, integer variables $\varRoomDayAgediff[r][d] \in \Znonneg$ are used to express the maximum age difference for room $r \in \rooms$ on day $d \in \days$.

The objective~\eqref{eq:room_obj} minimizes the sum of age differences in rooms, which is the only soft constraint related to the patient-to-room assignment.
\begin{equation}
  \min \quad \weightAgemix \sum_{r \in \rooms} \sum_{d \in \days} \varRoomDayAgediff[r][d] \label{eq:room_obj}
\end{equation}
Equations~\eqref{eq:room_once} ensure that exactly one room is assigned to every present patient.
Inequalities \eqref{eq:room_capacity} limit the number of patients staying in a certain room \edit{on a given day}, either by its capacity (if the gender matches) or by $0$.
Equations~\eqref{eq:room_onegender} \edit{enforce} that only one gender per room is allowed.
\begin{equation}
  \sum_{r \in \rooms} \varPatientRoom[p][r] = 1 \quad \forall p \in \presentPatients \label{eq:room_once}
\end{equation}
\begin{equation}
  \sum_{\substack{p \in \presentPatients \\ d \in \presentDays{p}, ~ \genderPatient{p} = g}} \varPatientRoom[p][r] \leq \roomCapacity{r} \varRoomDayGender[r][d][g] \quad  \forall r \in \rooms ~ \forall d \in \days ~ \forall g \in \genders \label{eq:room_capacity}
\end{equation}
\begin{equation}
  \sum_{g \in \genders} \varRoomDayGender[r][d][g] = 1 \quad \forall r \in \rooms ~ \forall d \in \days \label{eq:room_onegender}
\end{equation}
Inequalities \eqref{eq:room_age1} enforce $\varRoomDayAge$-variables to equal $1$ if a patient of a certain age group is present on a given day.
Using these variables, we can bound the age difference from below via inequalities \eqref{eq:room_age2}.
\begin{equation}
  \varRoomDayAge[r][d][\ageGroupPatient{p}] \geq \varPatientRoom[p][r] \quad \forall p \in \presentPatients ~ \forall d \in \presentDays{p} \label{eq:room_age1}
\end{equation}
\begin{equation}
  \varRoomDayAgediff[r][d] \geq |a-a'| \cdot ( \varRoomDayAge[r][d][a] + \varRoomDayAge[r][d][a'] - 1) \quad  \forall r \in \rooms ~ \forall d \in \days ~ \forall a,a' \in \ageGroups \label{eq:room_age2}
\end{equation}

Preliminary experiments showed that this problem is difficult to solve for MIP solvers.
A careful analysis of the linear programming relaxation revealed that the gender constraints can easily be ignored when the integer condition on variables is relaxed.
To see this, consider an assignment of patients to rooms by means of a vector $\varPatientRoom \in \{0,1\}^{\presentPatients \times \rooms}$ that respects the total room capacity but mixes patients of different genders.
Now, by setting $\varRoomDayGender[r][d][g] \in [0,1]$ according to the gender distribution (of gender $g$ in room $r$ on day $d$), we can easily extend it to a feasible solution \edit{of} the relaxation, which would be infeasible for the original MIP.
Hence, the MIP solver will have to resolve this problem by enumeration (or other techniques), which slows down the solution process.

For this reason, we implemented the same model using CP, and solved it using OR-Tools~\cite{ORTOOLS}, which was much faster in finding feasible solutions than Gurobi~\cite{GUROBI}.
However, \edit{we observed} that good performance \edit{requires} at least two threads, which we believe is due to the fact that OR-Tools was designed to run multiple search techniques in parallel.
Similar to the first subproblem, we use a callback mechanism to collect multiple feasible solutions for further processing.
The best solutions are added to the solution pool and progress to the patient-to-OT assignment.

\subsubsection{Patient-to-OT assignment}
\label{sec_approach_decomposition_theater}

Once the CP found a feasible patient-to-room assignment, we solve one MIP per day to assign the admitted patients to OTs.
Since these problems are independent for different days, we solve them in parallel again using the two threads that were previously used for the CP.
Since the OT assignment is independent of the rooms, we can use the found assignment for all patient-to-room-assignment solutions of a given admission solution.

\DeclareDocumentCommand\presentPatientsDay{m}{\ensuremath{\mathcal{P}^{\text{present}}_{#1}}}
\DeclareDocumentCommand\varPatientTheater{oo}{\ensuremath{\IfValueTF{#1}{x_{#1,#2}}{x}}}
\DeclareDocumentCommand\varSurgeonTheater{oo}{\ensuremath{\IfValueTF{#1}{y_{#1,#2}}{y}}}

Let $\presentPatientsDay{d^\star} \subseteq \presentPatients$ denote the patients with surgery on day $d^\star$.
Clearly, we can skip all days $d^\star$ with $\presentPatientsDay{d^\star} = \emptyset$.

Similar to the patient admission problem, we use variables $\varOpened[o] \in \{0,1\}$ to indicate whether operating theater $o \in \theaters$ is opened on day $d^\star$ ($\varOpened[o] = 1$) or not ($\varOpened[o] = 0$).
In addition, variables $\varPatientTheater[p][o] \in \{0,1\}$ state whether patient $p \in \presentPatientsDay{d^\star}$ is assigned to OT $o \in \theaters$ ($\varPatientTheater[p][o] = 1$) or not ($\varPatientTheater[p][o] = 0$).
Variables $\varSurgeonTheater[u][o] \in \{0,1\}$ indicate whether surgeon $u \in \surgeons$ is operating in OT $o \in \theaters$ on day $d^\star$ ($\varSurgeonTheater[u][o] = 1$) or not ($\varSurgeonTheater[u][o] = 0$).

The objective is to minimize the weighted sum of the number of opened OTs and the number of OTs assigned to each surgeon~\eqref{eq:ot_obj}.
Note that the second term does not reflect the number of surgeons \emph{transfers} between OTs as stated in the IHTC, but it is off by a constant that depends on the number of surgeons actually working on day $d^\star$.
\begin{equation}
  \min \quad \weightTheater \sum_{o \in \theaters} \varOpened[o] + \weightSurgeon \sum_{u \in \surgeons} \sum_{o \in \theaters} \varSurgeonTheater[u][o] \label{eq:ot_obj}
\end{equation}
Equations~\eqref{eq:ot_once} enforce that each patient has surgery in exactly one OT.
Additionally, we use the disaggregated formulation in~\eqref{eq:ot_cap1} and \eqref{eq:ot_cap2} to model that the OT capacity is respected and the variables representing open OTs are set correctly.
\begin{equation}
  \sum_{o \in \theaters} \varPatientTheater[p][o] = 1 \quad \forall p \in \presentPatientsDay{d^\star} \label{eq:ot_once}
\end{equation}
\begin{alignat}{7}
  \varOpened[o] &\geq \varPatientTheater[p][o] &\quad& \forall p \in \presentPatientsDay{d^\star} ~ \forall o \in \theaters \label{eq:ot_cap1}\\
  \sum_{p \in \presentPatientsDay{d^\star}} \durationSurgery{p} \varPatientTheater[p][o] &\leq \theaterCapacity{o}{d^\star} \varOpened[o] &\quad& \forall o \in \theaters\label{eq:ot_cap2}
\end{alignat}
Finally, inequalities~\eqref{eq:ot_surgeon} force $\varSurgeonTheater$-variables to~$1$ if the corresponding surgeon carries out at least one surgery in a specific OT on day $d^\star$.
\begin{equation}
  \varPatientTheater[p][o] \leq \varSurgeonTheater[\surgeonOf{p}][o] \quad \forall o \in \theaters ~ \forall p \in \presentPatientsDay{d^\star} \label{eq:ot_surgeon}
\end{equation}
The patient-to-OT assignment is solved with Gurobi~\cite{GUROBI}.
The optimal solution is combined with the respective patient admission and patient-to-room assignment solution to be further processed in the nurse-to-room assignment.

\subsubsection{Nurse-to-room assignment}
\label{sec_approach_decomposition_nurse}

Lastly, the fourth thread waits for feasible solutions (having an associated patient admission day schedule, patient-to-room, and patient-to-OT assignments) in order to add a nurse-to-room assignment.
Since finding an optimal nurse assignment turned out to be computationally difficult and many partial solutions arise from our decomposition approach, we need to evaluate each of them efficiently.
Therefore, we implemented \edit{an} SA approach for finding good heuristic nurse-to-room assignments.
Since we run the SA only on one thread and need to do this for many partial solutions, the SA is run once per solution.
The heuristic solution will be used as a warm start for the exact nurse-to-room assignment in Phase 3 (presented in \cref{sec_approach_exactnurse}).
The algorithm follows the standard outline of the SA algorithm, as outlined in~\cite{sa}.

\DeclareDocumentCommand\SAobj{m}{\ensuremath{\mathop{obj}(#1)}}
\DeclareDocumentCommand\SAbestsol{}{\ensuremath{x^\star}}
\DeclareDocumentCommand\SAiterStall{}{\ensuremath{\mathop{iter\_wo\_improve}}}
\DeclareDocumentCommand\SAmaxNoImprove{}{\ensuremath{\mathop{MaxNoImprove}}}

The nurse-to-room assignment is only affected by soft constraints for a fixed patient admission schedule and patient-to-room assignment, which means all solutions are feasible and no constraints need to be checked.
Thus, suitable neighborhood moves can be evaluated \edit{efficiently} using delta evaluations, i.e., only evaluating the effect of a potential change on the objective function.

We made the following design choices for the application in this context:
\begin{description}
\item[Solution representation:]
  A solution for the nurse-to-room assignment $\mathcal{X}$ is defined by the assignment of a nurse $n$ to each room $r$, day $d$ and shift $s$, i.e., $\mathcal{X}: \rooms \times \days \times \shifts \to \nurses \cup \{0\}$, where $0$ indicates no nurse.
\item[Construction heuristic:]
  The initial solution $\mathcal{X}^0$ is constructed in a greedy manner by iterating \edit{over} nurses and their work shifts, assigning unassigned rooms as long as the workload \edit{stays within its limits}.
  If there are unassigned rooms in shifts remaining after this initial step, the remaining rooms are iteratively assigned to the available nurse with the lowest workload in the given shift, resulting in the lowest workload excess possible.
\item[Temperature:]
  The temperature is cooled (using a factor $\Gamma = {0.999}$) in every iteration and initialized using the fitness \SAobj{\mathcal{X}^0} of the initial solution $\mathcal{X}^0$.
  We set the start temperature so that we accept solutions that are \SI{5}{\percent} worse than the initial solution $\mathcal{X}^0$ with a probability of 0.5. 
  This approach is suggested in~\cite{Ropke2006} to adapt the starting temperature to instance-specific values based on the initial solution.
  Thus, the initial temperature $T^0$ is set according to \eqref{eq:sa_t0}.
  \begin{equation}
    T^0 = \frac{1.05 \SAobj{\mathcal{X}^0} - \SAobj{\mathcal{X}^0}}{-\ln(0.5)}\label{eq:sa_t0}
  \end{equation}
\item[Neighbourhood operator:]
  The neighborhood operator picks a random shift and room and replaces the current nurse with a new nurse.
  The new nurse is determined by evaluating the difference in objective functions for all available nurses.
  The assignment is randomized but gives higher weight to nurses with better performance. 

  For a random room, day $d$, and shift $s$, we want to exchange the nurse. 
  We evaluate all available nurses $n \in \nursesShift{d}{s}$, i.e, all nurses with $\nurseCapacity{n}{d}{s} > 0$, for their change in the objective function ($\Phi_n$).
  We normalize all $\Phi_n$ to the interval $[0.02,1.02]$ with 0.02 being the worst solution and 1.02 being the best using \edit{Eq. \eqref{eq:sa_norm}}.
  \begin{equation}
    \Phi'_n = \frac{ (\Phi_n - \min_{i \in \nursesShift{d}{s} }{\{{\Phi_i}}\}) }{ (\max_{i \in \nursesShift{d}{s} }{\{\Phi_i\}}-\min_{i \in \nursesShift{d}{s} }{\{\Phi_i}\}) } + 0.02.
    \label{eq:sa_norm}
  \end{equation}
  We added the 0.02 to give a low probability to the worst-performing nurse, which otherwise would have 0 probability.
  This is to make sure that we can reach all solutions in the solution space.
  The probability of a nurse being chosen is then \edit{according to Eq. \eqref{eq:sa_weight}}.
  \begin{equation}
    Prob_n = \frac{\Phi'_n}{\sum_{i \in \nursesShift{d}{s} }\Phi'_i}.
    \label{eq:sa_weight}
  \end{equation}
  We chose this simple neighborhood operator over more complex neighborhood operators to exploit very fast evaluation of changes.
  More complicated neighborhoods, such as considering several nurses at the same time, would make this evaluation potentially more time-consuming.
\end{description}

\begin{algorithm}[H]
  \caption{Simulated annealing for heuristic nurse-to-room assignment}
  \begin{algorithmic}[1]
    \State Initialize with $\mathcal{X} \gets \mathcal{X}^0, T \gets T^0$ (according to \eqref{eq:sa_t0}).
    \State $\mathcal{X}^* \gets \mathcal{X}$
    \While{time not exceeded \textbf{and} $\SAiterStall < \SAmaxNoImprove$}
      \State Select room $r$ and shift $(d,s)$ uniformly at random.
      \State Get current nurse $n^{curr} \gets \mathcal{X}(r,d,s)$ 
      \State Calculate change in objective function $\Phi_{n} \forall n \in \mathcal{N}_{d,s}$ 
      \State Normalize $\Phi_{n}$ for all $n \in \mathcal{N}_{d,s}$ according to~\eqref{eq:sa_norm}.
      \State Select new nurse $n^{new}$ randomly based on probability $Prob_n$ as in~\eqref{eq:sa_weight}.
      \State $\mathcal{X}^{'} \gets $ Switch nurses $n^{curr}$ and $n^{new}$, i.e, $\mathcal{X}^{'}(r,d,s) \gets n^{new}$.

    \State Increase $\SAiterStall$ by $1$.
      \If{\SAobj{\mathcal{X}^{'}} $<$ \SAobj{\mathcal{X}^*}}
        \State $\mathcal{X}^* \gets \mathcal{X}^{'} $ \Comment{New best solution}
        \State Reset $\SAiterStall$ to $0$.
    \EndIf
      \If{\SAobj{\mathcal{X}^{'}} $<$ \SAobj{\mathcal{X}}}
        \State $\mathcal{X} \gets \mathcal{X}^{'}$ \Comment{Improving solution}
      \ElsIf{Random$(0,1) < \exp^{(-({obj}(\mathcal{X}^{'})-{obj}(\mathcal{X})) / T)}$}
        \State $\mathcal{X} \gets \mathcal{X}^{'}$ \Comment{Accepting worse solution}
      \EndIf
      \State $T \gets \Gamma \cdot T$ \Comment{Cooling}
    \EndWhile
    \State \Return $\mathcal{X}^*$
    \label{alg:sa}
  \end{algorithmic}
\end{algorithm}

\Cref{alg:sa} states the steps of our SA algorithm.
The calculation of differences in objective functions $\Phi_n$ for each potential nurse and the overall fitness calculation $\SAobj{\mathcal{X}}$ of a solution $\mathcal{X}$ consists of evaluating the three soft-constraints linked to the nurse assignment.
These are the excess workload of nurses, the continuity of care (i.e.\ the number of nurses assigned per patient), and using a lower skill than needed for a patient.
For mathematical formulations of these, we refer to \cref{sec_approach_exactnurse}.

During the SA, we keep track of the current workload per shift for each nurse, how often each nurse is assigned to each patient, and the sum of skill mismatches per nurse and shift.
In case a neighborhood move is accepted, the \edit{workload, assignment and skill mismatch penalities} are updated. 
Keeping \edit{these values updated} allows us to calculate the difference in objective functions $\Phi_n$ of assigning a new nurse $n$ to and removing another nurse from a room in a given shift, by just calculating the differences in excess workload, skill mismatch, and continuity for this individual room and shift.
To account for different weights of the soft constraints, we also use the weights $\weightWorkload$, $\weightCoC$, and $\weightSkill$ to calculate $\Phi_n$ as a weighted sum.

After termination of the \edit{SA} algorithm due to either time limit or a given number of iterations without improvement, the best computed nurse-to-room assignment is combined with the other subproblem solutions, resulting in a complete feasible solution to the original planning problem.
We further try to improve the solution quality in Phase~3.

\subsection{Phase~3: Finalization of the nurse-to-room assignment}
\label{sec_approach_exactnurse}

Once Phase~2 of the solution approach is complete or reached its time limit, we process the solutions that are complete, i.e, that have solution values for all four subproblems, and process them in order of increasing total cost.
The goal of Phase~3 is to \edit{reduce} the nurse-to-room assignment costs further by using an exact MIP approach for the nurse-to-room assignment compared to the heuristic in Phase~2.

In contrast to the MIP for patient-to-OT assignment and even for patient admission, this MIP is quite hard to solve exactly within a few minutes.
The patient admission MIP is not necessarily smaller, but the solver was able to produce several good solutions early on without large gaps to optimality.
We tried several approaches for the nurse-to-room assignment, such as Benders' decomposition~\cite{Benders62}, but without much success.
The best approach turned out to be the complete MIP as stated in~\eqref{eq:nurse_obj}--\eqref{eq:nurse_excess}.

\DeclareDocumentCommand\varNurseDayShiftRoom{oooo}{\ensuremath{\IfValueTF{#4}{x_{#1,#2,#3,#4}}{x}}}
\DeclareDocumentCommand\varNursePatient{oo}{\ensuremath{\IfValueTF{#2}{y_{#1,#2}}{y}}}
\DeclareDocumentCommand\varExcess{ooo}{\ensuremath{\IfValueTF{#3}{\varepsilon_{#1,#2,#3}}{\varepsilon}}}

The MIP uses decision variables $\varNurseDayShiftRoom[n][d][s][r] \in \{0,1\}$ to indicate whether nurse $n \in \nurses$ cares for patients in room $r \in \rooms$ during shift $s \in \shifts$ on day $d \in \days$ ($\varNurseDayShiftRoom[n][d][s][r] = 1$) or not ($\varNurseDayShiftRoom[n][d][s][r] = 0$).
Note that if nurse $n$ is not working during shift $s$ on day $d$, then $\varNurseDayShiftRoom[n][d][s][r]$ is a parameter equal to $0$.
Variables $\varNursePatient[n][p] \in \{0,1\}$ state whether nurse $n \in N$ cares for patient $p \in \presentPatients$ at least once ($\varNursePatient[n][p] = 1$) or never ($\varNursePatient[n][p] = 0$).
The variables $\varExcess[n][d][s] \in \Rnonneg$ measure the excess workload of nurse $n \in \nurses$ for day $d \in \days$ during shift $s \in \shifts$.

The objective is to minimize the weighted sum of costs for skill level mismatches, continuity of care, and excess workload~\eqref{eq:nurse_obj}.
\begin{multline}
  \min \ \weightSkill \sum_{n \in \nurses} \sum_{p \in \presentPatients} \sum_{\delta=0}^{\lengthStay{p}-1} \sum_{s \in \shifts} \max\{0, \skillNurse{n} - \skillPatient{p}{\delta}{s} \} \varNurseDayShiftRoom[n][d(p) + \delta][s][\roomOf{p}] \\
  + \weightCoC \sum_{n \in \nurses} \sum_{p \in \presentPatients} \varNursePatient[n][p]
  + \weightWorkload \sum_{n \in \nurses} \sum_{d \in \days} \sum_{s \in \shifts} \varExcess[n][d][s]\label{eq:nurse_obj}
\end{multline}
Equations~\eqref{eq:nurse_onenurse} ensure that there is a nurse in every room with at least one patient.
\begin{equation}
  \sum_{n \in \nurses} \varNurseDayShiftRoom[n][d][s][r] = 1 \quad \forall d \in \days ~ \forall s \in \shifts ~ \forall r \in \rooms : \exists p \in \presentPatientsDay{d}: r = \roomOf{p}\label{eq:nurse_onenurse}
\end{equation}
To keep track of continuity of care, inequality~\eqref{eq:nurse_coc} enforces $y_{n,p} = 1$ if nurse $n$ cares for patient $p$ at least once, where $M_{n,s} \leq |\presentDays{p}| \cdot |\shifts|$ denotes the number of pairs $(d,s) \in \presentDays{p} \times \shifts$ for which $\varNurseDayShiftRoom[n][d][s][r(p)]$ is a variable (and not a parameter).
\begin{equation}
  \sum_{d \in \presentDays{p}} \sum_{s \in \shifts} \varNurseDayShiftRoom[n][d][s][r(p)] \leq M_{n,s} \varNursePatient[n][p] \quad \forall n \in \nurses ~ \forall p \in \presentPatients\label{eq:nurse_coc}
\end{equation}
Inequalities~\eqref{eq:nurse_excess} calculate the excess workload per nurse and shift based on the given assignments.
\begin{equation}
  \varExcess[n][d][s] + \nurseCapacity{n}{d}{s} \geq \sum_{p \in \presentPatientsDay{d}} \durationPatient{p}{d - d(p)}{s} \varNurseDayShiftRoom[n][d][s][r(p)] \quad \forall n \in \nurses ~ \forall d \in \days ~ \forall s \in \shifts\label{eq:nurse_excess}
\end{equation}
In our experiments, we observed that primal/dual simplex methods \edit{require significant computational time} to solve the initial LP relaxation.
Hence, we parameterize Gurobi to use only the barrier method.

See \cref{fig:solution_approach} for the division among threads and passing of solutions.
If an optimal solution can be found within the time limit, then a thread \edit{selects} another solution and tries to improve upon the nurse-to-room assignment found by the heuristic.
The algorithm described so far \edit{is referred to} as the \emph{default algorithm} for the remainder of the paper.

\subsection{Rejection variant for patient admission}
\label{sec_rejection}

As already mentioned in \cref{sec_approach_carebounds}, the decomposition into four subproblems makes it difficult to consider the costs of later subproblems when admitting patients in the first subproblem.
Despite the fact that it yields lower bounds on the optimum, the solution approach is heuristic in nature, i.e., there is no guarantee for finding an optimal solution even if we run it without a time limit.
Given this viewpoint, \edit{the first subproblem can be seen} as a method for finding many promising admission solutions that will be evaluated and completed in later subproblems.

In this section, we will describe a \emph{rejection variant} for patient admission that we developed after the competition, whose goal is to find even more solutions for the patient admission subproblem.
In this variant, the interface between patient admission \edit{and} patient-to-room assignment is slightly modified as described below. Besides that, the remainder of the algorithm is unchanged.

The main idea is to reject all patient admission solutions proposed by the MIP solver in its callback method, which forces it to continue searching.
Consider such a proposed admission solution, given by the set $\presentPatients$ of admitted patients and the admission day $d(p)$ for each patient $p \in \presentPatients$. Recall that $x_{p,d}$ determines the admission day of patient $p$ (if $x_{p,d}=1$) and $\pi_p$ tells us if a patient $p$ is postponed (if $x_{p,d}=1$), then the rejection of a solution can be modeled by inequality~\eqref{eq_admission_nogood}.
\begin{equation}
  \sum_{p \in \flexiblePatients \cap \presentPatients} \varAdmit[p][d(p)] + \sum_{p \in \flexiblePatients \setminus \presentPatients} \varPostpone[p] \leq |\flexiblePatients| - 1, 
  \label{eq_admission_nogood}
\end{equation}
Inequality~\eqref{eq_admission_nogood} forbids this exact assignment of admission days to this particular set of admitted patients.
Since the inequality cuts off the proposed solution, the MIP solver will continue searching alternative solutions, even considering worse ones.
However, we still consider this admission solution for patient-to-room assignment, although we have rejected it for the admission problem, since it is a feasible admission schedule.

Unfortunately, just adding inequalities~\eqref{eq_admission_nogood} will \edit{significantly} degrade the solver's performance.
The reason is that the internal solver heuristics are never successful in finding solutions, so, in particular, powerful improvement heuristics (that take a feasible solution as input) are not used.
Fortunately, this negative side effect can be mitigated as follows.
We introduce the additional variable $h \in [0,1]$, which has a very large penalty in the objective function.
This variable is added to~\eqref{eq_admission_nogood}, resulting in the modified constraint~\eqref{eq_admission_nogood_workaround}.
\begin{equation}
  \sum_{p \in \flexiblePatients \cap \presentPatients} \varAdmit[p][d(p)] + \sum_{p \in \flexiblePatients \setminus \presentPatients} \varPostpone[p] \leq |\flexiblePatients| - 1 + h, 
  \label{eq_admission_nogood_workaround}
\end{equation}
The interpretation is that every admission solution $(x,z,\pi,\varepsilon,\theta,h)$ with $h=0$ is cut off, while the solution $(x,z,\pi,\varepsilon,\theta,1)$ will be feasible, paying the large (but constant) penalty.
Note that for an integer solution, there is no incentive for the solver to pick a fractional $h$, i.e., $h$ is an implied integer variable.
The correctness of the approach follows from the fact that every solution that was already accepted for room-to-patient assignment satisfies $h=1$, meaning that all feasible solutions have to pay the same penalty.
Since the $h$-variable only appears in constraints~\eqref{eq_admission_nogood_workaround} that are only relevant for already rejected solutions, the solver will have no incentive to produce solutions with $h=1$, meaning that its run is almost identical to that for the model without the $h$-variables.
The only difference is that, whenever we cut off a solution $(x,z,\pi,\varepsilon,\theta,0)$, we can propose the solution $(x,z,\pi,\varepsilon,\theta,1)$ to the solver instead (which is not cut off).
This added solution gives the solver's heuristics a good starting point, e.g., to run neighborhood search heuristics.

This rejection method also allows for another idea, which turned out to be difficult to integrate into our implementation. In the second subproblem, we check the existence of a feasible patient-to-room assignment for a given admission solution.
Since this is rather time-intensive, it would be beneficial to parallelize this feasibility check.
However, in an ordinary multi-stage MIP approach, we have to decide on the feasibility of a MIP solution before letting the solver continue its branch-and-bound process.
For example, if we later add a constraint that is violated by a previously accepted solution, then the solver's behavior may be undefined.
With our rejection algorithm, this cannot happen, at least if such constraints only cut off solutions with $h = 0$, since these are infeasible (due to the rejection) anyhow.
Such an approach could be particularly useful when solving multiple patient-to-room assignments subproblems in parallel.
However, this would require a redesign of our code, and hence we did not test this approach in practice.

\section{Experimental results}
\label{sec_experiments}

\DeclareDocumentCommand\instance{m}{\texttt{#1}}
\DeclareDocumentCommand\better{m}{\textcolor{green!60!black}{#1}}
\DeclareDocumentCommand\worse{m}{\textcolor{red}{#1}}

In this section, we first present the experimental setup and the published problem instances of the IHTC~2024\footnote{\url{https://ihtc2024.github.io/}} in  \cref{sec_experiments_implementation_hardware_instances}.
\Cref{sec_experiments_competition} presents the results of our approach for the IHTC competition setup, while \cref{sec_experiments_extended} provides results for an extended runtime.
\Cref{sec_soft} investigates soft constraint costs for specific instances.
In \cref{sec_experiments_results_sa,sec_experiments_results_rejections}, we analyze the contribution of the exact nurse-to-room assignment and the alternative patient admission model based on rejections (from \cref{sec_rejection}), respectively.

\subsection{Implementation, hardware and instances}
\label{sec_experiments_implementation_hardware_instances}

The implementation of the solution approach is available on GitHub~\cite{TeamTwente}.
The code is written in Python and uses the MIP solver Gurobi~12~\cite{GUROBI} via an academic license, as well as the open-source CP solver from Google's OR-Tools~\cite{ORTOOLS}.

Compared to the submission code for the IHTC24, we \edit{made} minor changes to fix errors that we noticed when rerunning the experiments on all instances.
This includes a corrected weighting of the nurse selection in the neighborhood operator (see \cref{sec_approach_decomposition_nurse}).
The previous version of the code did not normalize the weights to achieve probabilities summing up to 1.0, with that only taking a subset of better-performing nurses into account (and not all of them).
Additionally, some thread timing issues were fixed.

Our solutions were produced using cluster nodes \edit{with} 16 Intel Xeon Silver 4314 CPUs with four cores each, running on \SI{2.4}{\giga\hertz}, and with \SI{264}{\giga\byte} RAM.
We ran multiple experiments in parallel, but reserved eight CPU cores for the four threads in order to reduce interference with other computations.
For each experiment, every problem instance was solved 10 times, and we report the average results.
This compensates for small random effects, e.g., those due to parallelism.

There are several sets of published instances.
Instances \texttt{iXX} refer to the public instance set of the competition used to determine the finalists.
Instances \texttt{mXX} were the hidden instances to determine the final ranking among the five finalists.
Instance sets \texttt{smallXX} and \texttt{lXX\_X} were published after the competition for additional cases and analyses.
The former are aimed at comparing exact methods (and thus smaller), while the latter are even larger than the competition instances.

\subsection{Results for competition setup}
\label{sec_experiments_competition}

We briefly experimented with different working (time) limits for the different components and concluded with the following for our competition submission:
\begin{itemize}
\item 
  We allow \SI{300}{\second} for Phases~1 and~2 together and another \SI{300}{\second} for Phase~3.
  Phase~1 itself has no working limit, but typically completes in less than \SI{30}{\second}.
  Consequently, the remaining time can be used for Phase~2.
\item 
  We run the CP for each patient-to-room assignment (\cref{sec_approach_decomposition_room}) for at most \SI{5}{\second} and then continue with patient-to-OT assignment.
\item
  The reduction $\varrho$ of the aggregated room capacity is increased by $1$ after 6 infeasible CPs for solutions of the current $\varrho$-value.
\item
  For the patient-to-OT assignment, we allow \SI{1}{\second} per admission day.
\item
  The SA approach (\cref{sec_approach_decomposition_nurse}) is stopped either after a given runtime limit of \SI{15}{\second} or after \num{5000} iterations without improvement.
\end{itemize}
Besides that, we intentionally did not tune parameters for specific instances to keep the approach robust.
We use the canonical seed value~$0$ to ensure reproducibility in the SA component.
However, \edit{due to the} use \edit{of} multiple threads that interact as well as time limits, the overall algorithm is not deterministic.

Like in the competition setup, we ran our solution approach with a time limit of \SI{10}{\minute}.
We compare the best-known solution from the competition website\footnote{\href{https://ihtc2024.github.io/}{https://ihtc2024.github.io}, retrieved on 1 October 2025} to our solutions.
The results are presented in the columns labeled as \SI{10}{\minute} in \cref{table_results_quality}.

During the competition session at the EURO~2025 conference, it became apparent that other successful submissions were entirely heuristic-based approaches or used MIP only as a means to generate good solutions.
In contrast to that, our patient admission MIP (see \cref{sec_approach_decomposition_admission}) provides lower bounds on the optimum of the overall planning problem.
Due to the computation of the lower bounds on care costs in Phase~1 (see \cref{sec_approach_carebounds}), we even incorporated cost information from later stages into the patient admission MIP.
Consequently, we are confident that our lower bounds are actually (among) the best ones that were established during the competition phase.
These bounds are presented in the second column (\texttt{Lower Bound}) of \cref{table_results_quality}.

It is apparent from the fifth column (\texttt{Gap}) that, for many instances, our produced solutions are significantly worse than the best-known solutions (column 3).
However, they are closer \edit{to the best known solution} than the objective value difference to the lower bound.
In our experiments, we observed that the care-cost lower bound (see \cref{sec_approach_carebounds}) does not approximate the actual nurse assignment costs very well.
Hence, we suspect that the optima typically lie closer to the best-known solution values than to the lower bounds.

Moreover, \edit{in} roughly half of the instances, the feasibility of the patient-to-room assignment is not a problem ($\varrho = 0$, see \cref{sec_approach_decomposition_admission}), while \edit{in} the other half, one or more reductions of the aggregated room capacity are needed.
For some instances, e.g., \instance{i13}, \instance{m01} and \instance{m04}, a large $\varrho$-value goes hand in hand with a large gap to the best known solution, while there are also examples where this is not the case (\instance{m02} and \instance{m03}).
A possible reason is that a reduction of the aggregated room capacity can lead to fewer admitted patients, which is most likely not necessary over the whole time period to achieve feasibility.
We believe that a non-heuristic feedback mechanism from (infeasible) patient-to-room assignments to the patient admission problem is needed to exclude infeasible solutions.

Therefore, we developed the algorithm variant described in \cref{sec_rejection} in order to cut off admission solutions for which no feasible patient-to-room assignment exists.
Cutting off multiple second-stage infeasible solutions at once is one of the central ideas in logic-based Benders' decomposition~\cite{HookerO03}.
However, due to the computational costs of our CP approach in the patient-to-room assignment subproblem, further research or a parallelized approach would be needed.
As outlined at the end of \cref{sec_rejection}, we were unable to integrate such an idea into our competition code.

Detailed costs per soft constraint for each instance are given in \cref{table_results_details_default,table_results_details_long} in~\ref{sec:appendix}.

\newcolumntype{R}[1]{>{\raggedleft\arraybackslash}p{#1}}

\begin{center}
  \setlength{\tabcolsep}{2pt}
  \footnotesize
  \begin{longtable}{%
p{0.08\textwidth}%
R{0.07\textwidth}%
R{0.09\textwidth}%
R{0.11\textwidth}R{0.05\textwidth}R{0.02\textwidth}%
R{0.07\textwidth}
R{0.11\textwidth}R{0.05\textwidth}R{0.02\textwidth}%
}%
    \caption{%
      Solution values after running our algorithm once for \SI{10}{\minute} and \SI{1}{\hour}, respectively, including the objective value, gap to the best known solution [\%] and reduction $\varrho$ of aggregated room capacity at which the solution was found.
      A number in brackets after the objective value indicates the number of runs in which no feasible solution was found.
      In this case, the averages are computed over the remaining runs.
      The second column indicates the lower bound on the optimum derived from the \SI{1}{\hour} run, and the third column the best known solution value from the competition website~\cite{IHTC24}.
      The seventh column shows the improvement from the \SI{10}{\minute} to the \SI{1}{\hour} run [\%].
    }
    \label{table_results_quality} \\
\toprule
    Instance & Lower & Best sol & \multicolumn{3}{c}{Default \SI{10}{\minute}} & $\searrow$ & \multicolumn{3}{c}{Default \SI{1}{\hour}} \\
\cmidrule(r){4-6} \cmidrule(r){8-10}
          & bound & Obj. & Obj.\,[fails] & Gap & $\varrho$ & [\%] & Obj.\,[fails] & Gap & $\varrho$ \\
\midrule
\endfirsthead

\multicolumn{4}{c}{{Table \thetable\ (continued)}}\\
\toprule
    Instance & Lower & Best sol & \multicolumn{3}{c}{Default \SI{10}{\minute}} & $\searrow$ & \multicolumn{3}{c}{Default \SI{1}{\hour}} \\
\cmidrule(r){4-6} \cmidrule(r){8-10}
          & bound & Obj. & Obj.\,[fails] & Gap & $\varrho$ & [\%] & Obj.\,[fails] & Gap & $\varrho$ \\
\midrule
\endhead

\bottomrule
\endfoot

% python3 table_quality.py i01 i02 i03 i04 i05 i06 i07 i08 i09 i10 i11 i12 i13 i14 i15 i16 i17 i18 i19 i20 i21 i22 i23 i24 i25 i26 i27 i28 i29 i30
    \instance{i01} & \num{3700} & \num{3842} & \num{3886}\phantom{\,[0]} & $1.1$ & $0.0$ & $\better{0.0}$ & \num{3885}\phantom{\,[0]} & $1.1$ & $0.0$ \\
    \instance{i02} & \num{1049} & \num{1264} & \num{1375}\phantom{\,[0]} & $8.8$ & $0.3$ & $\worse{-0.1}$ & \num{1377}\phantom{\,[0]} & $8.9$ & $0.0$ \\
    \instance{i03} & \num{10335} & \num{10490} & \num{10536}\phantom{\,[0]} & $0.4$ & $1.2$ & $\better{0.0}$ & \num{10534}\phantom{\,[0]} & $0.4$ & $0.0$ \\
    \instance{i04} & \num{1564} & \num{1884} & \num{1898}\phantom{\,[0]} & $0.8$ & $1.8$ & $\better{0.0}$ & \num{1897}\phantom{\,[0]} & $0.7$ & $1.7$ \\
    \instance{i05} & \num{12642} & \num{12760} & \num{12807}\phantom{\,[0]} & $0.4$ & $0.0$ & $\worse{-0.0}$ & \num{12810}\phantom{\,[0]} & $0.4$ & $0.0$ \\
    \instance{i06} & \num{10576} & \num{10671} & \num{10680}\phantom{\,[0]} & $0.1$ & $0.0$ & $0.0$ & \num{10679}\phantom{\,[0]} & $0.1$ & $0.0$ \\
    \instance{i07} & \num{3490} & \num{5026} & \num{5410}\phantom{\,[0]} & $7.6$ & $5.1$ & $\better{0.7}$ & \num{5370}\phantom{\,[0]} & $6.9$ & $5.7$ \\
    \instance{i08} & \num{4733} & \num{6291} & \num{6414}\phantom{\,[0]} & $2.0$ & $0.9$ & $\better{1.2}$ & \num{6340}\phantom{\,[0]} & $0.8$ & $0.1$ \\
    \instance{i09} & \num{5448} & \num{6682} & \num{6937}\phantom{\,[0]} & $3.8$ & $0.9$ & $\better{1.1}$ & \num{6860}\phantom{\,[0]} & $2.7$ & $1.0$ \\
    \instance{i10} & \num{18240} & \num{20820} & \num{20881}\phantom{\,[0]} & $0.3$ & $0.4$ & $0.0$ & \num{20880}\phantom{\,[0]} & $0.3$ & $1.3$ \\
    \instance{i11} & \num{25811} & \num{25938} & \num{25962}\phantom{\,[0]} & $0.1$ & $0.0$ & $\better{0.0}$ & \num{25956}\phantom{\,[0]} & $0.1$ & $0.0$ \\
    \instance{i12} & \num{9385} & \num{12430} & \num{12932}\phantom{\,[0]} & $4.0$ & $0.6$ & $\better{0.5}$ & \num{12873}\phantom{\,[0]} & $3.6$ & $1.4$ \\
    \instance{i13} & \num{15264} & \num{17328} & \num{20815}\phantom{\,[0]} & $20.1$ & $2.4$ & $\better{3.1}$ & \num{20162}\phantom{\,[0]} & $16.4$ & $2.0$ \\
    \instance{i14} & \num{7997} & \num{9746} & \num{10266}\phantom{\,[0]} & $5.3$ & $0.0$ & $\better{4.4}$ & \num{9810}\phantom{\,[0]} & $0.7$ & $3.5$ \\
    \instance{i15} & \num{9405} & \num{12486} & \num{14306}\phantom{\,[0]} & $14.6$ & $3.8$ & $\better{8.2}$ & \num{13128}\phantom{\,[0]} & $5.1$ & $5.1$ \\
    \instance{i16} & \num{8678} & \num{10139} & \num{10527}\phantom{\,[0]} & $3.8$ & $1.0$ & $\worse{-0.4}$ & \num{10568}\phantom{\,[0]} & $4.2$ & $0.9$ \\
    \instance{i17} & \num{30575} & \num{40535} & \num{49116}\phantom{\,[0]} & $21.2$ & $1.5$ & $\better{1.5}$ & \num{48358}\phantom{\,[0]} & $19.3$ & $2.0$ \\
    \instance{i18} & \num{37246} & \num{37660} & \num{37682}\phantom{\,[0]} & $0.1$ & $1.6$ & $\better{0.1}$ & \num{37632}\phantom{\,[0]} & $\better{-0.1}$ & $1.8$ \\
    \instance{i19} & \num{36590} & \num{44587} & \num{48061}\phantom{\,[0]} & $7.8$ & $0.0$ & $\worse{-0.2}$ & \num{48152}\phantom{\,[0]} & $8.0$ & $0.0$ \\
    \instance{i20} & \num{26006} & \num{29098} & \num{29748}\phantom{\,[0]} & $2.2$ & $1.0$ & $\better{0.7}$ & \num{29543}\phantom{\,[0]} & $1.5$ & $0.6$ \\
    \instance{i21} & \num{18510} & \num{24703} & \num{27656}\phantom{\,[0]} & $12.0$ & $1.1$ & $\better{1.8}$ & \num{27150}\phantom{\,[0]} & $9.9$ & $1.5$ \\
    \instance{i22} & \num{38228} & \num{47861} & \num{50879}\phantom{\,[0]} & $6.3$ & $2.0$ & $\better{1.5}$ & \num{50096}\phantom{\,[0]} & $4.7$ & $2.0$ \\
    \instance{i23} & \num{33600} & \num{37550} & \num{39819}\phantom{\,[0]} & $6.0$ & $0.0$ & $\better{0.8}$ & \num{39497}\phantom{\,[0]} & $5.2$ & $0.1$ \\
    \instance{i24} & \num{31952} & \num{33221} & \num{34214}\phantom{\,[0]} & $3.0$ & $0.0$ & $\better{0.9}$ & \num{33898}\phantom{\,[0]} & $2.0$ & $0.0$ \\
    \instance{i25} & \num{9106} & \num{11517} & \num{12700}\phantom{\,[0]} & $10.3$ & $0.0$ & $\better{1.0}$ & \num{12572}\phantom{\,[0]} & $9.2$ & $0.0$ \\
    \instance{i26} & \num{53768} & \num{64613} & \num{70426}\phantom{\,[0]} & $9.0$ & $1.2$ & $\better{0.8}$ & \num{69890}\phantom{\,[0]} & $8.2$ & $1.0$ \\
    \instance{i27} & \num{37302} & \num{51828} & \num{59948}\phantom{\,[0]} & $15.7$ & $0.4$ & $\better{1.5}$ & \num{59029}\phantom{\,[0]} & $13.9$ & $0.7$ \\
    \instance{i28} & \num{72732} & \num{75172} & \num{76429}\phantom{\,[0]} & $1.7$ & $0.0$ & $\better{0.6}$ & \num{75946}\phantom{\,[0]} & $1.0$ & $0.1$ \\
    \instance{i29} & \num{10164} & \num{12475} & \num{12890}\phantom{\,[0]} & $3.3$ & $2.0$ & $\worse{-0.3}$ & \num{12935}\phantom{\,[0]} & $3.7$ & $2.0$ \\
    \instance{i30} & \num{31697} & \num{37943} & \num{38798}\phantom{\,[0]} & $2.3$ & $0.5$ & $\better{1.8}$ & \num{38105}\phantom{\,[0]} & $0.4$ & $0.7$ \\

\midrule

% python3 table_quality.py m01 m02 m03 m04 m05 m06 m07 m08 m09 m10 m11 m12 m13 m14 m15 m16 m17 m18 m19 m20 m21 m22 m23 m24 m25 m26 m27 m28 m29 m30
    \instance{m01} & \num{2968} & \num{3384} & \num{9421}\phantom{\,[0]} & $178.4$ & $6.3$ & $\better{18.3}$ & \num{7695}\phantom{\,[0]} & $127.4$ & $9.0$ \\
    \instance{m02} & \num{12117} & \num{12211} & \num{12238}\phantom{\,[0]} & $0.2$ & $3.2$ & $\worse{-0.0}$ & \num{12239}\phantom{\,[0]} & $0.2$ & $0.3$ \\
    \instance{m03} & \num{6419} & \num{6697} & \num{6756}\phantom{\,[0]} & $0.9$ & $4.0$ & $0.0$ & \num{6756}\phantom{\,[0]} & $0.9$ & $4.0$ \\
    \instance{m04} & \num{2544} & \num{3345} & \num{10040}\phantom{\,[0]} & $200.1$ & $2.0$ & $\better{0.0}$ & \num{10037}\phantom{\,[0]} & $200.1$ & $2.0$ \\
    \instance{m05} & \num{11596} & \num{11956} & \num{11995}\phantom{\,[0]} & $0.3$ & $0.0$ & $0.0$ & \num{11994}\phantom{\,[0]} & $0.3$ & $0.0$ \\
    \instance{m06} & \num{26950} & \num{28250} & \num{28340}\phantom{\,[0]} & $0.3$ & $0.0$ & $\better{0.6}$ & \num{28180}\phantom{\,[0]} & $\better{-0.2}$ & $0.0$ \\
    \instance{m07} & \num{6657} & \num{7329} & \num{8138}\phantom{\,[0]} & $11.0$ & $1.0$ & $0.0$ & \num{8138}\phantom{\,[0]} & $11.0$ & $1.0$ \\
    \instance{m08} & \num{13070} & \num{14976} & \num{15838}\phantom{\,[0]} & $5.8$ & $0.0$ & $\better{4.9}$ & \num{15064}\phantom{\,[0]} & $0.6$ & $0.0$ \\
    \instance{m09} & \num{31272} & \num{32967} & \num{32969}\phantom{\,[0]} & $0.0$ & $0.4$ & $0.0$ & \num{32967}\phantom{\,[0]} & $0.0$ & $1.7$ \\
    \instance{m10} & \num{24884} & \num{26027} & \num{26149}\phantom{\,[0]} & $0.5$ & $2.5$ & $\better{0.2}$ & \num{26101}\phantom{\,[0]} & $0.3$ & $3.7$ \\
    \instance{m11} & \num{33146} & \num{35030} & \num{35495}\phantom{\,[0]} & $1.3$ & $0.0$ & $\better{1.2}$ & \num{35055}\phantom{\,[0]} & $0.1$ & $0.0$ \\
    \instance{m12} & \num{9338} & \num{11681} & \num{12463}\phantom{\,[0]} & $6.7$ & $1.0$ & $\better{1.9}$ & \num{12221}\phantom{\,[0]} & $4.6$ & $1.0$ \\
    \instance{m13} & \num{30459} & \num{31189} & \num{31616}\phantom{\,[0]} & $1.4$ & $0.0$ & $\better{0.9}$ & \num{31334}\phantom{\,[0]} & $0.5$ & $0.2$ \\
    \instance{m14} & \num{12071} & \num{13368} & \num{13405}\phantom{\,[0]} & $0.3$ & $0.0$ & $\better{0.4}$ & \num{13358}\phantom{\,[0]} & $\better{-0.1}$ & $0.0$ \\
    \instance{m15} & \num{24146} & \num{28727} & \num{30306}\phantom{\,[0]} & $5.5$ & $2.7$ & $\better{3.9}$ & \num{29121}\phantom{\,[0]} & $1.4$ & $2.4$ \\
    \instance{m16} & \num{12040} & \num{17271} & \num{18592}\phantom{\,[0]} & $7.7$ & $0.0$ & $\better{0.8}$ & \num{18439}\phantom{\,[0]} & $6.8$ & $0.1$ \\
    \instance{m17} & \num{20765} & \num{22542} & \num{22631}\phantom{\,[0]} & $0.4$ & $0.0$ & $\better{0.1}$ & \num{22615}\phantom{\,[0]} & $0.3$ & $0.7$ \\
    \instance{m18} & \num{5785} & \num{7817} & \num{8504}\phantom{\,[0]} & $8.8$ & $0.0$ & $\worse{-0.0}$ & \num{8507}\phantom{\,[0]} & $8.8$ & $0.5$ \\
    \instance{m19} & \num{19761} & \num{21790} & \num{23298}\phantom{\,[0]} & $6.9$ & $2.0$ & $\better{3.9}$ & \num{22395}\phantom{\,[0]} & $2.8$ & $0.3$ \\
    \instance{m20} & \num{3890} & \num{5335} & \num{5751}\phantom{\,[0]} & $7.8$ & $0.0$ & $\better{3.9}$ & \num{5526}\phantom{\,[0]} & $3.6$ & $0.8$ \\
    \instance{m21} & \num{20858} & \num{25170} & \num{25686}\phantom{\,[0]} & $2.0$ & $1.0$ & $\better{4.9}$ & \num{24433}\phantom{\,[0]} & $\better{-2.9}$ & $1.0$ \\
    \instance{m22} & \num{17060} & \num{21330} & \num{24011}\phantom{\,[0]} & $12.6$ & $0.0$ & $\better{0.9}$ & \num{23785}\phantom{\,[0]} & $11.5$ & $0.0$ \\
    \instance{m23} & \num{15150} & \num{18301} & \num{19243}\phantom{\,[0]} & $5.1$ & $0.0$ & $\worse{-0.4}$ & \num{19312}\phantom{\,[0]} & $5.5$ & $0.0$ \\
    \instance{m24} & \num{28042} & \num{30873} & \num{31754}\phantom{\,[0]} & $2.9$ & $0.0$ & $\better{1.2}$ & \num{31383}\phantom{\,[0]} & $1.7$ & $0.0$ \\
    \instance{m25} & \num{32088} & \num{33985} & \num{34293}\phantom{\,[0]} & $0.9$ & $0.5$ & $\worse{-0.6}$ & \num{34498}\phantom{\,[0]} & $1.5$ & $0.9$ \\
    \instance{m26} & \num{63987} & \num{69730} & \num{73529}\,[1] & $5.4$ & $0.0$ & $\worse{-0.1}$ & \num{73573}\phantom{\,[0]} & $5.5$ & $0.0$ \\
    \instance{m27} & \num{17725} & \num{28028} & \num{36129}\phantom{\,[0]} & $28.9$ & $0.0$ & $\better{1.6}$ & \num{35542}\phantom{\,[0]} & $26.8$ & $0.0$ \\
    \instance{m28} & \num{34075} & \num{45913} & \num{54416}\phantom{\,[0]} & $18.5$ & $0.0$ & $\better{2.0}$ & \num{53310}\,[3] & $16.1$ & $0.0$ \\
    \instance{m29} & \num{44740} & \num{48082} & \num{48808}\phantom{\,[0]} & $1.5$ & $0.0$ & $\worse{-20.0}$ & \num{58579}\phantom{\,[0]} & $21.8$ & $0.0$ \\
    \instance{m30} & \num{26970} & \num{31649} & \num{32981}\,[1] & $4.2$ & $0.0$ & $-0.0$ & \num{32982}\,[5] & $4.2$ & $0.6$ \\

\midrule

% python3 table_quality.py small01 small02 small03 small04 small05 small06 small07 small08 small09
    \instance{small01} & \num{233} & $\infty$ & \num{242}\phantom{\,[0]} &  & $0.0$ & $0.0$ & \num{242}\phantom{\,[0]} &  & $0.0$ \\
    \instance{small02} & \num{690} & $\infty$ & $\infty$\phantom{{\,[0]}} &  &  &  & $\infty$\phantom{{\,[0]}} &  &  \\
    \instance{small03} & \num{3860} & $\infty$ & \num{4015}\phantom{\,[0]} &  & $0.0$ & $0.0$ & \num{4015}\phantom{\,[0]} &  & $0.0$ \\
    \instance{small04} & \num{1992} & $\infty$ & \num{5242}\phantom{\,[0]} &  & $2.0$ & $0.0$ & \num{5242}\phantom{\,[0]} &  & $2.0$ \\
    \instance{small05} & \num{3215} & $\infty$ & $\infty$\phantom{{\,[0]}} &  &  &  & $\infty$\phantom{{\,[0]}} &  &  \\
    \instance{small06} & \num{2534} & $\infty$ & $\infty$\phantom{{\,[0]}} &  &  &  & $\infty$\phantom{{\,[0]}} &  &  \\
    \instance{small07} & \num{2797} & $\infty$ & \num{4001}\phantom{\,[0]} &  & $1.0$ & $0.0$ & \num{4001}\phantom{\,[0]} &  & $1.0$ \\
    \instance{small08} & \num{6085} & $\infty$ & \num{6575}\phantom{\,[0]} &  & $0.0$ & $0.0$ & \num{6575}\phantom{\,[0]} &  & $0.0$ \\
    \instance{small09} & \num{5285} & $\infty$ & \num{6450}\phantom{\,[0]} &  & $1.0$ & $-0.0$ & \num{6450}\phantom{\,[0]} &  & $1.0$ \\

\midrule

% python3 table_quality.py l35_1 l35_2 l35_3 l35_4 l35_5 l35_6 l42_1 l42_2 l42_3 l42_4 l42_5 l42_6 l49_1 l49_2 l49_3 l49_4 l49_5 l49_6 l56_1 l56_2 l56_3 l56_4 l56_5 l56_6
    \instance{l35\_1} & \num{6621} & $\infty$ & \num{7935}\phantom{\,[0]} &  & $1.0$ & $\better{0.2}$ & \num{7916}\phantom{\,[0]} &  & $1.0$ \\
    \instance{l35\_2} & \num{30517} & $\infty$ & \num{30832}\phantom{\,[0]} &  & $0.0$ & $\better{0.4}$ & \num{30706}\phantom{\,[0]} &  & $0.0$ \\
    \instance{l35\_3} & \num{19098} & $\infty$ & \num{28575}\phantom{\,[0]} &  & $1.0$ & $\better{0.4}$ & \num{28472}\phantom{\,[0]} &  & $1.1$ \\
    \instance{l35\_4} & \num{23060} & $\infty$ & \num{32708}\phantom{\,[0]} &  & $0.0$ & $\better{1.1}$ & \num{32347}\phantom{\,[0]} &  & $0.0$ \\
    \instance{l35\_5} & \num{60567} & $\infty$ & \num{194390}\phantom{\,[0]} &  & $1.0$ & $\better{54.5}$ & \num{88524}\phantom{\,[0]} &  & $2.0$ \\
    \instance{l35\_6} & \num{51445} & $\infty$ & $\infty$\phantom{{\,[0]}} &  &  &  & $\infty$\phantom{{\,[0]}} &  &  \\
    \instance{l42\_1} & \num{13100} & $\infty$ & \num{20712}\phantom{\,[0]} &  & $2.9$ & $\better{1.7}$ & \num{20362}\phantom{\,[0]} &  & $2.8$ \\
    \instance{l42\_2} & \num{8095} & $\infty$ & \num{11657}\phantom{\,[0]} &  & $0.1$ & $\better{0.1}$ & \num{11645}\phantom{\,[0]} &  & $0.4$ \\
    \instance{l42\_3} & \num{70554} & $\infty$ & \num{87741}\phantom{\,[0]} &  & $2.3$ & $\better{4.8}$ & \num{83516}\phantom{\,[0]} &  & $2.0$ \\
    \instance{l42\_4} & \num{126131} & $\infty$ & \num{129668}\phantom{\,[0]} &  & $0.0$ & $\better{0.6}$ & \num{128908}\phantom{\,[0]} &  & $0.0$ \\
    \instance{l42\_5} & \num{120305} & $\infty$ & \num{144285}\phantom{\,[0]} &  & $0.0$ & $\better{1.0}$ & \num{142823}\phantom{\,[0]} &  & $0.5$ \\
    \instance{l42\_6} & \num{89792} & $\infty$ & \num{107124}\phantom{\,[0]} &  & $0.0$ & $\better{1.6}$ & \num{105387}\phantom{\,[0]} &  & $0.3$ \\
    \instance{l49\_1} & \num{19051} & $\infty$ & \num{25119}\phantom{\,[0]} &  & $2.0$ & $\better{0.6}$ & \num{24970}\phantom{\,[0]} &  & $2.0$ \\
    \instance{l49\_2} & \num{82164} & $\infty$ & \num{91002}\phantom{\,[0]} &  & $1.0$ & $\better{1.2}$ & \num{89870}\phantom{\,[0]} &  & $1.4$ \\
    \instance{l49\_3} & \num{60007} & $\infty$ & \num{82350}\phantom{\,[0]} &  & $2.0$ & $\better{1.6}$ & \num{81031}\phantom{\,[0]} &  & $3.0$ \\
    \instance{l49\_4} & \num{101492} & $\infty$ & \num{114625}\phantom{\,[0]} &  & $0.0$ & $\better{3.0}$ & \num{111210}\phantom{\,[0]} &  & $0.1$ \\
    \instance{l49\_5} & \num{64641} & $\infty$ & \num{87323}\,[8] &  & $0.5$ & $\better{1.5}$ & \num{85987}\,[5] &  & $4.4$ \\
    \instance{l49\_6} & \num{110208} & $\infty$ & \num{130029}\,[6] &  & $1.0$ & $\better{1.3}$ & \num{128394}\,[2] &  & $4.9$ \\
    \instance{l56\_1} & \num{25775} & $\infty$ & \num{35065}\phantom{\,[0]} &  & $3.0$ & $\better{3.0}$ & \num{34010}\phantom{\,[0]} &  & $3.0$ \\
    \instance{l56\_2} & \num{50028} & $\infty$ & \num{58277}\,[1] &  & $2.2$ & $\better{1.7}$ & \num{57305}\phantom{\,[0]} &  & $2.0$ \\
    \instance{l56\_3} & \num{101624} & $\infty$ & \num{117385}\phantom{\,[0]} &  & $0.8$ & $\better{0.6}$ & \num{116689}\phantom{\,[0]} &  & $0.7$ \\
    \instance{l56\_4} & \num{152835} & $\infty$ & \num{216614}\phantom{\,[0]} &  & $1.0$ & $\better{16.0}$ & \num{181908}\phantom{\,[0]} &  & $1.0$ \\
    \instance{l56\_5} & \num{93126} & $\infty$ & \num{110791}\phantom{\,[0]} &  & $0.0$ & $\better{1.3}$ & \num{109365}\phantom{\,[0]} &  & $1.0$ \\
    \instance{l56\_6} & \num{182912} & $\infty$ & $\infty$\phantom{{\,[0]}} &  &  &  & \num{236250}\,[4] &  & $5.2$ \\

  \end{longtable}
\end{center}

\subsection{Results for extended running time}
\label{sec_experiments_extended}

Although our implementation is tailored towards the \SI{10}{\minute} time limit, it is interesting to investigate whether the results improve with a longer runtime.
To this end, we scaled the time limits mentioned in \cref{sec_experiments_competition} by a factor of 6 to a total runtime of 1\,h.
The results are presented in the columns labeled as \SI{1}{\hour} in \cref{table_results_quality}.
Moreover, the relative improvement from \SI{10}{\minute} to \SI{1}{\hour} is shown in the seventh column ($\searrow$).

The first observation from the results is that running the algorithm longer does not always give a benefit.
In fact, for several instances (highlighted in red), the found solutions were worse, but at maximum $\SI{0.5}{\percent}$ (\instance{m25}).
This can happen due to the non-deterministic character \edit{of the algorithm}.
However, \edit{larger instances clearly benefit from }having more runtime.
In particular, for instances \instance{m06}, \instance{m14} and \instance{m21}, we could even improve upon the best known solution value.

Finally, it is worth mentioning that the \SI{1}{\hour} runs for \instance{m01} found a much better solution than the \SI{10}{\minute} runs despite a much larger $\varrho$-value of $9$.

\subsection{Analysis of soft constraint violations}
\label{sec_soft}

For several instances with a gap of at least \SI{10}{\percent} to the best known solution, we present the detailed costs per soft constraint for our solution approach (with a time limit of 10 minutes) and the best known solution from the competition website in \cref{table_results_details_bad}.
The difference in performance has one of two reasons in these instances.
Either our approach leaves more patients unscheduled, i.e., the patient admission \edit{struggles} to find solutions with many patients that lead to feasible room assignments, or the nurse assignment has high costs (continuity of care, excess workload, and skill mismatch).

\Cref{table_results_details_bad} also reveals that, in our approach, delays after the release day are lower in most of the solutions (except instance \instance{m27}).
This is because the delay costs are included in the first subproblem on patient admission, and low-cost admission solutions are prioritized for consideration by later subproblems.
For the other soft constraints, the approaches lead to somewhat similar costs.

\begin{center}
  \setlength{\tabcolsep}{2pt}
  \footnotesize%
  \begin{longtable}{%
    p{0.09\textwidth}%
    R{0.08\textwidth}%
    R{0.092\textwidth}%
    R{0.10\textwidth}%
    R{0.08\textwidth}%
    R{0.08\textwidth}%
    R{0.08\textwidth}%
    R{0.07\textwidth}%
    R{0.09\textwidth}%
    R{0.09\textwidth}%
  }
    \caption{%
      Detailed soft constraint costs for instances with a gap of at least \SI{10}{\percent} for the best known solution of the competition (\textit{Best}) and our default algorithm (\textit{Default}) with 10~minutes time limit, averaged over 10~runs.
      We also include the objective weights ($\omega$).
      Bold values indicate significantly higher cost in our solution.
      (CoC = Continuity of care, Unsch.\ = unscheduled optional patients, Exc.\ WL = excess workload, Open OTs = opened operating theaters, Delay = delayed patients, Age Mix = age mix in rooms, Skill match = underqualified nurses, surgeon transfer between operating theaters).
    }
    \label{table_results_details_bad} \\
\toprule
    Instance & Sol./$\omega$ & CoC & Unsch. & Exc. WL & Open OTs & Delay & Age Mix & Skill Match & Surgeon Transfer \\
\midrule
\endfirsthead

\multicolumn{4}{c}{{Table \thetable\ (continued)}}\\
\toprule
    Instance & Solution & CoC & Unsch. & Exc. WL & Open OTs & Delay & Age Mix & Skill Match & Surgeon Transfer \\ 
\midrule
\endhead

\bottomrule
\endfoot

    \instance{i13} & Best & $2865.0$ & $5500.0$ & $290.0$ & $300.0$ & $7815.0$ & $200.0$ & $318.0$ & $40.0$ \\
      & Default & $2616.5$ & $\textbf{10350.0}$ & $258.8$ & $277.0$ & $6868.5$ & $178.0$ & $240.4$ & $26.0$ \\
      & $\omega$ & $5$ & $500$ & $1$ & $10$ & $15$ & $5$ & $1$ & $10$ \\
\midrule
    \instance{i15} & Best & $3765.0$ & $2450.0$ & $50.0$ & $360.0$ & $5120.0$ & $91.0$ & $635.0$ & $15.0$ \\
      & Default & $\textbf{5009.5}$ & $2695.0$ & $483.0$ & $341.0$ & $4514.0$ & $38.5$ & $1204.0$ & $20.5$ \\
      & $\omega$ & $5$ & $350$ & $10$ & $10$ & $10$ & $1$ & $5$ & $5$ \\
\midrule
    \instance{i17} & Best & $8540.0$ & $14000.0$ & $270.0$ & $2220.0$ & $12040.0$ & $310.0$ & $3040.0$ & $115.0$ \\
      & Default & $\textbf{12261.0}$ & $13750.0$ & $\textbf{2392.0}$ & $2085.0$ & $10269.5$ & $605.5$ & $\textbf{7672.0}$ & $81.5$ \\
      & $\omega$ & $5$ & $500$ & $10$ & $30$ & $5$ & $5$ & $10$ & $5$ \\
\midrule
    \instance{i27} & Best & $3062.0$ & $24000.0$ & $380.0$ & $3240.0$ & $16205.0$ & $331.0$ & $4240.0$ & $370.0$ \\
      & Default & $\textbf{4485.2}$ & $21550.0$ & $\textbf{3927.0}$ & $3195.0$ & $14238.5$ & $211.8$ & $\textbf{12087.0}$ & $253.0$ \\
      & $\omega$ & $1$ & $500$ & $10$ & $30$ & $5$ & $1$ & $10$ & $10$ \\
\midrule
    \instance{m01} & Best & $362.0$ & $250.0$ & $170.0$ & $560.0$ & $1590.0$ & $43.0$ & $409.0$ & $0.0$ \\
      & Default & $151.9$ & $\textbf{7675.0}$ & $0.0$ & $344.0$ & $1146.0$ & $0.0$ & $104.3$ & $0.0$ \\
      & $\omega$ & $1$ & $250$ & $5$ & $40$ & $15$ & $1$ & $1$ & $10$ \\
\midrule
    \instance{m04} & Best & $339.0$ & $0.0$ & $181.0$ & $640.0$ & $2020.0$ & $165.0$ & $0.0$ & $0.0$ \\
      & Default & $209.1$ & $\textbf{7350.0}$ & $78.3$ & $480.0$ & $1840.0$ & $80.5$ & $2.0$ & $0.0$ \\
      & $\omega$ & $1$ & $350$ & $1$ & $40$ & $10$ & $5$ & $10$ & $10$ \\
\midrule
    \instance{m07} & Best & $393.0$ & $4800.0$ & $20.0$ & $540.0$ & $1355.0$ & $41.0$ & $180.0$ & $0.0$ \\
      & Default & $364.2$ & $\textbf{5700.0}$ & $0.0$ & $510.0$ & $1370.0$ & $32.0$ & $162.2$ & $0.0$ \\
      & $\omega$ & $1$ & $300$ & $10$ & $30$ & $5$ & $1$ & $1$ & $5$ \\
\midrule
    \instance{m22} & Best & $5715.0$ & $8400.0$ & $100.0$ & $1060.0$ & $5240.0$ & $165.0$ & $630.0$ & $20.0$ \\
      & Default & $\textbf{7780.0}$ & $8400.0$ & $660.5$ & $986.0$ & $3718.0$ & $296.0$ & $\textbf{2155.0}$ & $15.5$ \\
      & $\omega$ & $5$ & $400$ & $5$ & $20$ & $10$ & $5$ & $10$ & $1$ \\
\midrule
    \instance{m27} & Best & $6325.0$ & $13800.0$ & $80.0$ & $1680.0$ & $3690.0$ & $360.0$ & $2075.0$ & $18.0$ \\
      & Default & $\textbf{12530.5}$ & $4995.0$ & $\textbf{2061.0}$ & $2160.0$ & $\textbf{6144.0}$ & $422.5$ & $\textbf{7799.0}$ & $17.2$ \\
      & $\omega$ & $5$ & $150$ & $10$ & $40$ & $15$ & $5$ & $5$ & $1$ \\
\midrule
    \instance{m28} & Best & $13920.0$ & $8400.0$ & $170.0$ & $4880.0$ & $16440.0$ & $58.0$ & $1565.0$ & $480.0$ \\
      & Default & $\textbf{23030.0}$ & $7875.0$ & $\textbf{3179.5}$ & $4500.0$ & $12766.0$ & $33.9$ & $\textbf{2673.0}$ & $359.0$ \\
      & $\omega$ & $5$ & $350$ & $5$ & $40$ & $10$ & $1$ & $5$ & $10$ \\

  \end{longtable}
\end{center}

\subsection{Contribution of exact nurse-to-room assignment}
\label{sec_experiments_results_sa}

In this section, we analyze the effect of the additional exact nurse-to-room assignment in Phase~3 in comparison to the heuristic result at the end of Phase 2 (i.e., heuristic nurse-to-room assignment with SA).

\Cref{table_results_extra1} shows, in addition to the objective values of the best known solution and the 10-minute default approach, the relative improvement on the nurse-to-room assignment cost from the end of Phase~2 to the end of Phase 3 (column \emph{Nurse} $\searrow$).
The results for the additional instances are given in Appendix \cref{table_results_extra2}.

Phase~3 improves the overall solution quality for all instances, for some only slightly, while for some the improvement is significant.
There is no clear pattern with respect to the instance size, e.g., for the large instance \instance{m27} the improvement is only \SI{3.3}{\percent} while for the large instance \instance{m29} the improvement is \SI{66.7}{\percent}.
It is noteworthy, that for the instances \instance{i17}, \instance{i27}, \instance{m22} and \instance{m27} that were pointed out with high nurse-to-room assignment costs in \cref{sec_soft}, the improvement was below \SI{5}{\percent} (exception is instance \instance{m28} with \SI{31.6}{\percent}).
For 29 of the 60 instances in \cref{table_results_extra1}, the improvement was below \SI{5}{\percent} (and for 39 instances below \SI{10}{\percent}), indicating that SA provides a good solution in the sense that the performance is relatively close to the MIP approach.
\edit{In general, }the results highlight the difficulty of the nurse-to-room assignment, since the MIP is struggling to improve the heuristic solution even when given \SI{50}{\percent} of the overall runtime.
In future research, there is potential for improving the SA method further to provide better start solutions or replacing the MIP entirely.

\begin{center}
  \setlength{\tabcolsep}{2pt}
  \footnotesize%
  \begin{longtable}{%
p{0.08\textwidth}%
R{0.08\textwidth}%
R{0.09\textwidth}%
R{0.11\textwidth}R{0.10\textwidth}R{0.07\textwidth}%
R{0.07\textwidth}%
R{0.10\textwidth}%
R{0.07\textwidth}%
}%
    \caption{%
      Results for our default algorithm and the variant described in \cref{sec_rejection} for \SI{10}{\minute}, each \edit{run} 10 times per competition instance.
      See \cref{table_results_extra2} in \ref{sec:appendix} for the remaining instances.
      Similar to \cref{table_results_quality}, lower bounds and best known solutions are shown.
      For each of the algorithms, the (averaged) objective value (column obj) of the best found solution and the number of produced admission solutions are reported.
      A number in brackets after the objective value indicates the number of runs in which no feasible solution was found.
      In this case, the averages are computed over the remaining runs.
      In addition, column nurse\,$\searrow$ shows the relative improvement of the nurse costs (sum of continuity of care, excess workload, and skill match) from the end of Phase~2 to the end of Phase~3.
      The results are discussed in \cref{sec_experiments_results_sa,sec_experiments_results_rejections}.
    }
    \label{table_results_extra1} \\
\toprule
    Instance & Lower & Best sol & \multicolumn{3}{c}{Default \SI{10}{\minute}} & $\searrow$ & \multicolumn{2}{c}{Rejection \SI{10}{\minute}} \\
\cmidrule(r){4-6} \cmidrule(r){8-9} 
          & bound & Obj. & Nurse\,$\searrow$  & Obj.\,[fails] & \#sols & [\%] & Obj.\,[fails] & \#sols \\
\midrule
\endfirsthead

\multicolumn{4}{c}{{Table \thetable\ (continued)}}\\
\toprule
    Instance & Lower & Best sol & \multicolumn{3}{c}{Default \SI{10}{\minute}} & $\searrow$ & \multicolumn{2}{c}{Rejection \SI{10}{\minute}} \\
\cmidrule(r){4-6} \cmidrule(r){8-9} 
          & bound & Obj. & Nurse\,$\searrow$  & Obj.\,[fails] & \#sols & [\%] & Obj.\,[fails] & \#sols \\
\midrule
\endhead

\bottomrule
\endfoot

% python3 table_extras.py i01 i02 i03 i04 i05 i06 i07 i08 i09 i10 i11 i12 i13 i14 i15 i16 i17 i18 i19 i20 i21 i22 i23 i24 i25 i26 i27 i28 i29 i30
    \instance{i01} & \num{3700} & \num{3842}      & $0.8\,\%$ & \num{3886}\phantom{\,[0]} & \num{44.5}    & $\better{0.8}$ & \num{3855}\phantom{\,[0]} & \num{929.1} \\
    \instance{i02} & \num{1049} & \num{1264}      & $15.4\,\%$ & \num{1375}\phantom{\,[0]} & \num{110.0}    & $\better{1.1}$ & \num{1360}\phantom{\,[0]} & \num{1972.0} \\
    \instance{i03} & \num{10335} & \num{10490}      & $2.6\,\%$ & \num{10536}\phantom{\,[0]} & \num{51.0}    & $\better{0.2}$ & \num{10520}\phantom{\,[0]} & \num{913.4} \\
    \instance{i04} & \num{1564} & \num{1884}      & $20.1\,\%$ & \num{1898}\phantom{\,[0]} & \num{139.0}    & $\worse{-0.2}$ & \num{1902}\phantom{\,[0]} & \num{1080.7} \\
    \instance{i05} & \num{12642} & \num{12760}      & $17.0\,\%$ & \num{12807}\phantom{\,[0]} & \num{16.8}    & $\better{0.2}$ & \num{12787}\phantom{\,[0]} & \num{615.6} \\
    \instance{i06} & \num{10576} & \num{10671}      & $12.4\,\%$ & \num{10680}\phantom{\,[0]} & \num{87.0}    & $\better{0.0}$ & \num{10676}\phantom{\,[0]} & \num{379.9} \\
    \instance{i07} & \num{3490} & \num{5026}      & $10.0\,\%$ & \num{5410}\phantom{\,[0]} & \num{110.4}    & $\worse{-0.8}$ & \num{5456}\phantom{\,[0]} & \num{443.4} \\
    \instance{i08} & \num{4733} & \num{6291}      & $7.1\,\%$ & \num{6414}\phantom{\,[0]} & \num{102.8}    & $\better{1.6}$ & \num{6310}\phantom{\,[0]} & \num{338.4} \\
    \instance{i09} & \num{5448} & \num{6682}      & $7.5\,\%$ & \num{6937}\phantom{\,[0]} & \num{67.7}    & $\better{0.5}$ & \num{6902}\phantom{\,[0]} & \num{1384.1} \\
    \instance{i10} & \num{18240} & \num{20820}      & $15.6\,\%$ & \num{20881}\phantom{\,[0]} & \num{73.4}    & $\better{0.2}$ & \num{20844}\phantom{\,[0]} & \num{892.6} \\
    \instance{i11} & \num{25811} & \num{25938}      & $12.7\,\%$ & \num{25962}\phantom{\,[0]} & \num{44.9}    & $\worse{-0.0}$ & \num{25964}\phantom{\,[0]} & \num{351.2} \\
    \instance{i12} & \num{9385} & \num{12430}      & $9.0\,\%$ & \num{12932}\phantom{\,[0]} & \num{113.6}    & $\better{0.1}$ & \num{12913}\phantom{\,[0]} & \num{484.7} \\
    \instance{i13} & \num{15264} & \num{17328}      & $10.1\,\%$ & \num{20815}\phantom{\,[0]} & \num{215.9}    & $\better{5.0}$ & \num{19768}\phantom{\,[0]} & \num{1488.0} \\
    \instance{i14} & \num{7997} & \num{9746}      & $4.8\,\%$ & \num{10266}\phantom{\,[0]} & \num{27.0}    & $\worse{-0.3}$ & \num{10293}\phantom{\,[0]} & \num{383.0} \\
    \instance{i15} & \num{9405} & \num{12486}      & $11.8\,\%$ & \num{14306}\phantom{\,[0]} & \num{100.8}    & $\better{2.4}$ & \num{13965}\phantom{\,[0]} & \num{676.4} \\
    \instance{i16} & \num{8678} & \num{10139}      & $4.4\,\%$ & \num{10527}\phantom{\,[0]} & \num{86.8}    & $\worse{-0.4}$ & \num{10566}\phantom{\,[0]} & \num{760.3} \\
    \instance{i17} & \num{30575} & \num{40535}      & $2.1\,\%$ & \num{49116}\phantom{\,[0]} & \num{187.6}    & $\worse{-5.3}$ & \num{51742}\phantom{\,[0]} & \num{234.4} \\
    \instance{i18} & \num{37246} & \num{37660}      & $17.6\,\%$ & \num{37682}\phantom{\,[0]} & \num{86.2}    & $\worse{-0.0}$ & \num{37684}\phantom{\,[0]} & \num{473.6} \\
    \instance{i19} & \num{36590} & \num{44587}      & $1.2\,\%$ & \num{48061}\phantom{\,[0]} & \num{52.9}    & $\worse{-1.0}$ & \num{48534}\phantom{\,[0]} & \num{156.9} \\
    \instance{i20} & \num{26006} & \num{29098}      & $27.1\,\%$ & \num{29748}\phantom{\,[0]} & \num{110.2}    & $\worse{-0.0}$ & \num{29758}\phantom{\,[0]} & \num{887.2} \\
    \instance{i21} & \num{18510} & \num{24703}      & $0.5\,\%$ & \num{27656}\phantom{\,[0]} & \num{133.8}    & $\worse{-0.8}$ & \num{27886}\phantom{\,[0]} & \num{389.6} \\
    \instance{i22} & \num{38228} & \num{47861}      & $1.4\,\%$ & \num{50879}\phantom{\,[0]} & \num{210.2}    & $\worse{-4.9}$ & \num{53383}\phantom{\,[0]} & \num{312.3} \\
    \instance{i23} & \num{33600} & \num{37550}      & $4.2\,\%$ & \num{39819}\phantom{\,[0]} & \num{49.0}    & $\worse{-1.4}$ & \num{40380}\phantom{\,[0]} & \num{211.5} \\
    \instance{i24} & \num{31952} & \num{33221}      & $9.0\,\%$ & \num{34214}\phantom{\,[0]} & \num{63.0}    & $\worse{-0.6}$ & \num{34435}\phantom{\,[0]} & \num{105.5} \\
    \instance{i25} & \num{9106} & \num{11517}      & $0.4\,\%$ & \num{12700}\phantom{\,[0]} & \num{52.5}    & $\worse{-0.7}$ & \num{12791}\phantom{\,[0]} & \num{399.1} \\
    \instance{i26} & \num{53768} & \num{64613}      & $3.2\,\%$ & \num{70426}\phantom{\,[0]} & \num{193.9}    & $\worse{-2.0}$ & \num{71846}\phantom{\,[0]} & \num{369.7} \\
    \instance{i27} & \num{37302} & \num{51828}      & $4.1\,\%$ & \num{59948}\phantom{\,[0]} & \num{125.2}    & $\worse{-6.9}$ & \num{64096}\phantom{\,[0]} & \num{129.8} \\
    \instance{i28} & \num{72732} & \num{75172}      & $0.7\,\%$ & \num{76429}\phantom{\,[0]} & \num{31.6}    & $\worse{-0.3}$ & \num{76674}\phantom{\,[0]} & \num{192.2} \\
    \instance{i29} & \num{10164} & \num{12475}      & $0.5\,\%$ & \num{12890}\phantom{\,[0]} & \num{224.9}    & $\worse{-2.7}$ & \num{13241}\phantom{\,[0]} & \num{857.8} \\
    \instance{i30} & \num{31697} & \num{37943}      & $1.1\,\%$ & \num{38798}\phantom{\,[0]} & \num{100.5}    & $\worse{-1.7}$ & \num{39460}\phantom{\,[0]} & \num{257.1} \\

\midrule

% python3 table_extras.py m01 m02 m03 m04 m05 m06 m07 m08 m09 m10 m11 m12 m13 m14 m15 m16 m17 m18 m19 m20 m21 m22 m23 m24 m25 m26 m27 m28 m29 m30
    \instance{m01} & \num{2968} & \num{3384}      & $0.1\,\%$ & \num{9421}\phantom{\,[0]} & \num{99.6}    & $\worse{-40.2}$ & \num{13205}\phantom{\,[0]} & \num{145.0} \\
    \instance{m02} & \num{12117} & \num{12211}      & $9.5\,\%$ & \num{12238}\phantom{\,[0]} & \num{72.4}    & $\better{0.0}$ & \num{12232}\phantom{\,[0]} & \num{472.4} \\
    \instance{m03} & \num{6419} & \num{6697}      & $5.1\,\%$ & \num{6756}\phantom{\,[0]} & \num{44.9}    & $\better{0.1}$ & \num{6750}\phantom{\,[0]} & \num{1413.3} \\
    \instance{m04} & \num{2544} & \num{3345}      & $3.5\,\%$ & \num{10040}\phantom{\,[0]} & \num{84.8}    & $\worse{-0.6}$ & \num{10103}\phantom{\,[0]} & \num{643.2} \\
    \instance{m05} & \num{11596} & \num{11956}      & $1.9\,\%$ & \num{11995}\phantom{\,[0]} & \num{63.6}    & $\worse{-0.0}$ & \num{12000}\phantom{\,[0]} & \num{534.7} \\
    \instance{m06} & \num{26950} & \num{28250}      & $11.9\,\%$ & \num{28340}\phantom{\,[0]} & \num{33.6}    & $\worse{-1.0}$ & \num{28638}\phantom{\,[0]} & \num{173.1} \\
    \instance{m07} & \num{6657} & \num{7329}      & $19.1\,\%$ & \num{8138}\phantom{\,[0]} & \num{158.1}    & $\worse{-4.8}$ & \num{8533}\phantom{\,[0]} & \num{2135.3} \\
    \instance{m08} & \num{13070} & \num{14976}      & $4.3\,\%$ & \num{15838}\phantom{\,[0]} & \num{33.7}    & $\worse{-0.4}$ & \num{15906}\phantom{\,[0]} & \num{176.1} \\
    \instance{m09} & \num{31272} & \num{32967}      & $20.5\,\%$ & \num{32969}\phantom{\,[0]} & \num{162.0}    & $\worse{-0.0}$ & \num{32970}\phantom{\,[0]} & \num{311.9} \\
    \instance{m10} & \num{24884} & \num{26027}      & $7.4\,\%$ & \num{26149}\phantom{\,[0]} & \num{103.6}    & $\better{0.0}$ & \num{26141}\phantom{\,[0]} & \num{455.8} \\
    \instance{m11} & \num{33146} & \num{35030}      & $3.2\,\%$ & \num{35495}\phantom{\,[0]} & \num{37.0}    & $\worse{-0.7}$ & \num{35745}\phantom{\,[0]} & \num{273.6} \\
    \instance{m12} & \num{9338} & \num{11681}      & $0.6\,\%$ & \num{12463}\phantom{\,[0]} & \num{133.0}    & $\worse{-1.8}$ & \num{12690}\phantom{\,[0]} & \num{568.2} \\
    \instance{m13} & \num{30459} & \num{31189}      & $2.7\,\%$ & \num{31616}\phantom{\,[0]} & \num{36.7}    & $\better{0.2}$ & \num{31544}\phantom{\,[0]} & \num{274.4} \\
    \instance{m14} & \num{12071} & \num{13368}      & $31.2\,\%$ & \num{13405}\phantom{\,[0]} & \num{43.0}    & $\worse{-0.2}$ & \num{13437}\phantom{\,[0]} & \num{314.2} \\
    \instance{m15} & \num{24146} & \num{28727}      & $0.4\,\%$ & \num{30306}\phantom{\,[0]} & \num{264.6}    & $\worse{-5.3}$ & \num{31907}\phantom{\,[0]} & \num{441.5} \\
    \instance{m16} & \num{12040} & \num{17271}      & $9.8\,\%$ & \num{18592}\phantom{\,[0]} & \num{38.0}    & $\worse{-1.1}$ & \num{18796}\phantom{\,[0]} & \num{344.6} \\
    \instance{m17} & \num{20765} & \num{22542}      & $2.7\,\%$ & \num{22631}\phantom{\,[0]} & \num{37.4}    & $\worse{-0.0}$ & \num{22633}\phantom{\,[0]} & \num{353.3} \\
    \instance{m18} & \num{5785} & \num{7817}      & $6.8\,\%$ & \num{8504}\phantom{\,[0]} & \num{40.6}    & $\worse{-1.4}$ & \num{8619}\phantom{\,[0]} & \num{387.9} \\
    \instance{m19} & \num{19761} & \num{21790}      & $0.5\,\%$ & \num{23298}\phantom{\,[0]} & \num{171.7}    & $\better{2.5}$ & \num{22707}\phantom{\,[0]} & \num{1218.8} \\
    \instance{m20} & \num{3890} & \num{5335}      & $3.9\,\%$ & \num{5751}\phantom{\,[0]} & \num{48.1}    & $\worse{-1.3}$ & \num{5825}\phantom{\,[0]} & \num{421.5} \\
    \instance{m21} & \num{20858} & \num{25170}      & $5.5\,\%$ & \num{25686}\phantom{\,[0]} & \num{88.8}    & $\better{0.0}$ & \num{25684}\phantom{\,[0]} & \num{620.5} \\
    \instance{m22} & \num{17060} & \num{21330}      & $0.7\,\%$ & \num{24011}\phantom{\,[0]} & \num{53.2}    & $\better{0.9}$ & \num{23794}\phantom{\,[0]} & \num{309.3} \\
    \instance{m23} & \num{15150} & \num{18301}      & $0.6\,\%$ & \num{19243}\phantom{\,[0]} & \num{70.7}    & $\worse{-1.6}$ & \num{19552}\phantom{\,[0]} & \num{351.4} \\
    \instance{m24} & \num{28042} & \num{30873}      & $40.2\,\%$ & \num{31754}\phantom{\,[0]} & \num{43.8}    & $\worse{-0.8}$ & \num{32022}\phantom{\,[0]} & \num{254.9} \\
    \instance{m25} & \num{32088} & \num{33985}      & $11.5\,\%$ & \num{34293}\phantom{\,[0]} & \num{131.8}    & $\worse{-1.2}$ & \num{34694}\phantom{\,[0]} & \num{511.3} \\
    \instance{m26} & \num{63987} & \num{69730}      & $47.8\,\%$ & \num{73529}\,[1] & \num{66.5}    & $\worse{-4.4}$ & \num{76738}\phantom{\,[0]} & \num{216.6} \\
    \instance{m27} & \num{17725} & \num{28028}      & $3.3\,\%$ & \num{36129}\phantom{\,[0]} & \num{41.4}    & $\worse{-0.6}$ & \num{36346}\phantom{\,[0]} & \num{273.5} \\
    \instance{m28} & \num{34075} & \num{45913}      & $31.6\,\%$ & \num{54416}\phantom{\,[0]} & \num{74.6}    & $\worse{-0.1}$ & \num{54465}\,[3] & \num{395.7} \\
    \instance{m29} & \num{44740} & \num{48082}      & $66.7\,\%$ & \num{48808}\phantom{\,[0]} & \num{43.0}    & $\worse{-0.7}$ & \num{49126}\phantom{\,[0]} & \num{220.7} \\
    \instance{m30} & \num{26970} & \num{31649}      & $41.9\,\%$ & \num{32981}\,[1] & \num{115.4}    & $\worse{-6.1}$ & \num{34980}\,[6] & \num{639.1} \\
    
  \end{longtable}
\end{center}

\subsection{Results for rejection variant}
\label{sec_experiments_results_rejections}

In this section, we investigate the performance of the rejection variant presented in Section \ref{sec_rejection}.
The columns reporting about the number of generated admission solutions (\texttt{\#sols}) in \cref{table_results_extra1} clearly show that the algorithm variant successfully generates more solutions than our default algorithm.

The impact on the quality of computed solutions is less clear.
While this larger number of solutions is advantageous for smaller instances, there is a clear performance degradation for larger instances.
The results for the remaining non-competition instances can be found in \cref{table_results_extra2} in \ref{sec:appendix}.

\section{Conclusions}
\label{sec_conclusions}

In this paper, we present our proposed solution approach to the IHTC 2024.
We decomposed the overall planning problem into several subproblems, while still using aggregated information from later subproblems.
Our solution approach runs on four threads in parallel, maintaining a pool of (partial) solutions along the way.
To solve the subproblem efficiently, we used MIP, CP, and SA.
Apart from feasible solutions, we also compute lower bounds for each instance.

From our results, we conclude that our approach has improvement potential, which hopefully stimulates further research.
We would like to point out some research directions and ideas that we believe are worth investigating in order to improve exact approaches for the IHTC.

First, our decomposition approach has no feedback mechanism.
In particular, the parameter $\varrho$ for reducing the total aggregated room capacity allocated during the patient admission can certainly be improved.
Ideally, the patient-to-room assignment algorithm for distributing patients to rooms \edit{should}, if infeasible, identify a reason for this infeasibility that can be communicated back to the patient admission subproblem.
An example may be a restricted set of patient admissions that, by themselves, already yield infeasibility.

Second, the exact nurse assignment (see \cref{sec_approach_exactnurse,sec_experiments_results_sa}) is difficult for MIP approaches.
The reason seems to be that the continuity-of-care requirements are orthogonal to the excess workload requirements.
The former aims at allocating few nurses to each patient (which implies allocation to other patients since they share a room), and an optimal allocation may easily exceed the workload of some nurses.
The latter aims at distributing nurses evenly over the rooms.
This, however, can easily be obtained by considering convex combinations of entirely different nurse-patient assignments (which are feasible \edit{only} for the linear programming relaxation).
If all of them have a low continuity-of-care penalty, averaging maintains this property, \edit{and obtains} an even workload distribution.
Preventing \edit{this in }the linear programming relaxation, \edit{e.g.,} via suitable cutting planes, is an open research problem.

Another research line is to improve the heuristic nurse-to-room assignment by developing more sophisticated neighborhood operators for the SA, which would \edit{decrease} the speed per iteration but could lead to overall better solutions.
Furthermore, the SA nurse-to-room assignment is currently only run once per solution.
To improve the performance, several runs with different starting points and seed values could lead to better solutions.

A final observation is that a multi-stage approach \edit{is} quite suitable for a competition that allows parallelization.
This is in contrast to monolithic MIP approaches that typically take not much advantage from parallelization, except for solving the initial linear relaxation using different methods.
For our method, parallelization of the different stages was a structural advantage.
In fact, we believe that parallelizing even further could pay off, first to improve the later subproblems' performance (in particular the CP solver), second to be able to treat more partial solutions, and third to allow computing dual feedback in the spirit of logic-based Benders' decomposition (see last paragraph of \cref{sec_experiments_results_rejections}).

\paragraph{Acknowledgments.}
The second and the fifth author acknowledge funding support from the Dutch Research Council (NWO) on grant OCENW.M20.151.
We thank anonymous reviewers for their valuable suggestions and constructive feedback that led to an improved manuscript.

\paragraph{Competing interests.}
After submission the second author became affiliated with Mosek ApS, which produces proprietary optimization software.
The remaining authors have no competing interests to declare.

\bibliographystyle{plainurl}
\bibliography{twente-algorithm}

\newpage

\appendix

\section{Detailed results}
\label{sec:appendix}

\Cref{table_results_details_default,table_results_details_long} present the disaggregated objective function for the runs with 10\,min and 1\,h, respectively, showing the cost per soft constraint.

\begin{center}
  \setlength{\tabcolsep}{2pt}
  \footnotesize
  \begin{longtable}{%
    p{0.09\textwidth}%
    R{0.08\textwidth}%
    R{0.10\textwidth}%
    R{0.08\textwidth}%
    R{0.08\textwidth}%
    R{0.08\textwidth}%
    R{0.07\textwidth}%
    R{0.08\textwidth}%
    R{0.09\textwidth}%
    R{0.09\textwidth}%
  }
    \caption{%
      Detailed average costs per instance and soft constraint for the 10\,min runs (CoC = Continuity of care, Unsch.\ = unscheduled optional patients, Exc.\ WL = excess workload, Open OTs = opened operating theaters, Delay = delayed patients, Age Mix = age mix in rooms, Skill match = underqualified nurses, surgeon transfer between operating theaters).
      Empty cells indicate that no solution was found.
    }%
    \label{table_results_details_default}\\

\toprule
Instance & CoC & Unsch. & Exc. WL & Open OTs & Delay & Age Mix & Skill Match & Surgeon Transfer \\ 
\midrule
\endfirsthead

\multicolumn{4}{c}{{Table \thetable\ (continued)}}\\
\toprule
Instance & CoC & Unsched. & Excess. WL & Open OTs & Delay & Age Mix & Skill Match & Surgeon Transfer \\ 
\midrule
\endhead

\bottomrule
\endfoot

% python3 table_soft.py default i01 i02 i03 i04 i05 i06 i07 i08 i09 i10 i11 i12 i13 i14 i15 i16 i17 i18 i19 i20 i21 i22 i23 i24 i25 i26 i27 i28 i29 i30
    \instance{i01} & $125.9$ & $2800.0$ & $19.0$ & $240.0$ & $470.0$ & $11.0$ & $220.0$ & $0.0$ \\
    \instance{i02} & $242.2$ & $0.0$ & $24.0$ & $240.0$ & $589.0$ & $34.5$ & $245.5$ & $0.0$ \\
    \instance{i03} & $650.5$ & $7600.0$ & $8.0$ & $400.0$ & $1725.0$ & $20.0$ & $132.5$ & $0.0$ \\
    \instance{i04} & $350.3$ & $0.0$ & $74.0$ & $280.0$ & $954.0$ & $42.0$ & $197.9$ & $0.0$ \\
    \instance{i05} & $346.3$ & $8500.0$ & $3.0$ & $480.0$ & $3427.0$ & $5.0$ & $46.0$ & $0.0$ \\
    \instance{i06} & $234.4$ & $9900.0$ & $0.0$ & $240.0$ & $250.0$ & $12.0$ & $43.1$ & $0.0$ \\
    \instance{i07} & $2428.5$ & $0.0$ & $205.5$ & $348.0$ & $1407.0$ & $78.5$ & $940.0$ & $2.6$ \\
    \instance{i08} & $1147.6$ & $200.0$ & $882.3$ & $1700.0$ & $2257.0$ & $162.0$ & $10.0$ & $55.0$ \\
    \instance{i09} & $2461.0$ & $0.0$ & $210.5$ & $422.0$ & $3441.0$ & $200.5$ & $163.2$ & $39.0$ \\
    \instance{i10} & $3083.5$ & $12600.0$ & $171.0$ & $460.0$ & $2613.0$ & $163.0$ & $1790.5$ & $0.0$ \\
    \instance{i11} & $325.0$ & $24500.0$ & $0.0$ & $570.0$ & $493.0$ & $2.1$ & $72.3$ & $0.0$ \\
    \instance{i12} & $3068.0$ & $5200.0$ & $863.5$ & $813.0$ & $1093.5$ & $15.1$ & $1859.0$ & $19.5$ \\
    \instance{i13} & $2616.5$ & $10350.0$ & $258.8$ & $277.0$ & $6868.5$ & $178.0$ & $240.4$ & $26.0$ \\
    \instance{i14} & $3946.5$ & $3850.0$ & $136.0$ & $1100.0$ & $839.0$ & $20.2$ & $367.0$ & $7.0$ \\
    \instance{i15} & $5009.5$ & $2695.0$ & $483.0$ & $341.0$ & $4514.0$ & $38.5$ & $1204.0$ & $20.5$ \\
    \instance{i16} & $853.5$ & $4050.0$ & $867.6$ & $1360.0$ & $2940.0$ & $70.5$ & $355.0$ & $30.5$ \\
    \instance{i17} & $12261.0$ & $13750.0$ & $2392.0$ & $2085.0$ & $10269.5$ & $605.5$ & $7672.0$ & $81.5$ \\
    \instance{i18} & $830.6$ & $33300.0$ & $2.0$ & $750.0$ & $2653.0$ & $6.5$ & $137.9$ & $2.0$ \\
    \instance{i19} & $11330.0$ & $16350.0$ & $2194.5$ & $2001.0$ & $12811.5$ & $92.1$ & $3074.0$ & $208.0$ \\
    \instance{i20} & $951.0$ & $21550.0$ & $285.0$ & $1376.0$ & $3394.5$ & $46.9$ & $2139.0$ & $5.8$ \\
    \instance{i21} & $9334.0$ & $5250.0$ & $834.5$ & $2755.0$ & $6125.0$ & $142.0$ & $3195.0$ & $20.7$ \\
    \instance{i22} & $13188.5$ & $26775.0$ & $1054.0$ & $1644.0$ & $6486.5$ & $167.1$ & $1538.1$ & $25.5$ \\
    \instance{i23} & $9960.0$ & $19800.0$ & $1718.0$ & $682.0$ & $7377.0$ & $66.8$ & $207.0$ & $8.1$ \\
    \instance{i24} & $2553.6$ & $28630.0$ & $68.0$ & $660.0$ & $1721.0$ & $5.9$ & $445.6$ & $130.0$ \\
    \instance{i25} & $5890.5$ & $2400.0$ & $623.7$ & $1026.0$ & $2033.5$ & $178.5$ & $468.4$ & $79.5$ \\
    \instance{i26} & $17428.5$ & $32650.0$ & $2659.0$ & $852.0$ & $15019.0$ & $233.5$ & $1487.3$ & $97.0$ \\
    \instance{i27} & $4485.2$ & $21550.0$ & $3927.0$ & $3195.0$ & $14238.5$ & $211.8$ & $12087.0$ & $253.0$ \\
    \instance{i28} & $6310.5$ & $63900.0$ & $944.3$ & $2560.0$ & $2235.0$ & $4.5$ & $421.5$ & $53.0$ \\
    \instance{i29} & $1866.1$ & $4200.0$ & $300.2$ & $660.0$ & $4606.0$ & $441.0$ & $696.1$ & $121.0$ \\
    \instance{i30} & $8836.5$ & $16980.0$ & $3105.3$ & $2728.0$ & $6276.0$ & $603.5$ & $250.0$ & $18.9$ \\

\midrule

% python3 table_soft.py default m01 m02 m03 m04 m05 m06 m07 m08 m09 m10 m11 m12 m13 m14 m15 m16 m17 m18 m19 m20 m21 m22 m23 m24 m25 m26 m27 m28 m29 m30
    \instance{m01} & $151.9$ & $7675.0$ & $0.0$ & $344.0$ & $1146.0$ & $0.0$ & $104.3$ & $0.0$ \\
    \instance{m02} & $238.8$ & $10200.0$ & $2.0$ & $700.0$ & $1020.0$ & $4.7$ & $72.0$ & $0.0$ \\
    \instance{m03} & $828.0$ & $5400.0$ & $37.0$ & $80.0$ & $335.0$ & $0.0$ & $76.0$ & $0.0$ \\
    \instance{m04} & $209.1$ & $7350.0$ & $78.3$ & $480.0$ & $1840.0$ & $80.5$ & $2.0$ & $0.0$ \\
    \instance{m05} & $340.2$ & $9600.0$ & $169.4$ & $750.0$ & $960.0$ & $25.0$ & $150.0$ & $0.0$ \\
    \instance{m06} & $3086.0$ & $20500.0$ & $14.0$ & $820.0$ & $3322.5$ & $15.0$ & $583.0$ & $0.0$ \\
    \instance{m07} & $364.2$ & $5700.0$ & $0.0$ & $510.0$ & $1370.0$ & $32.0$ & $162.2$ & $0.0$ \\
    \instance{m08} & $4971.5$ & $8575.0$ & $142.0$ & $480.0$ & $1061.0$ & $16.1$ & $552.0$ & $40.0$ \\
    \instance{m09} & $1212.1$ & $16000.0$ & $1256.6$ & $1170.0$ & $13098.0$ & $6.1$ & $216.0$ & $10.0$ \\
    \instance{m10} & $2641.5$ & $19775.0$ & $228.5$ & $1000.0$ & $1990.5$ & $31.5$ & $481.9$ & $0.0$ \\
    \instance{m11} & $5259.5$ & $17700.0$ & $614.2$ & $2200.0$ & $9615.0$ & $4.6$ & $67.0$ & $35.0$ \\
    \instance{m12} & $4878.5$ & $2625.0$ & $475.4$ & $478.0$ & $2830.0$ & $462.0$ & $680.9$ & $33.5$ \\
    \instance{m13} & $1305.1$ & $17500.0$ & $66.0$ & $486.0$ & $11895.0$ & $43.8$ & $276.2$ & $44.0$ \\
    \instance{m14} & $1076.1$ & $9100.0$ & $8.0$ & $636.0$ & $1852.0$ & $7.0$ & $694.5$ & $31.0$ \\
    \instance{m15} & $2006.8$ & $7300.0$ & $2429.7$ & $1440.0$ & $16270.0$ & $698.0$ & $48.0$ & $114.0$ \\
    \instance{m16} & $5539.5$ & $5250.0$ & $2632.5$ & $1680.0$ & $1734.0$ & $30.9$ & $1663.0$ & $62.5$ \\
    \instance{m17} & $1037.2$ & $17000.0$ & $1347.0$ & $660.0$ & $2263.0$ & $9.0$ & $309.0$ & $6.0$ \\
    \instance{m18} & $1403.0$ & $2250.0$ & $1498.6$ & $432.0$ & $2406.5$ & $448.0$ & $9.0$ & $56.5$ \\
    \instance{m19} & $1213.0$ & $13050.0$ & $141.4$ & $1170.0$ & $7198.0$ & $241.0$ & $217.2$ & $67.0$ \\
    \instance{m20} & $1588.4$ & $1200.0$ & $711.1$ & $422.0$ & $1492.5$ & $201.5$ & $90.5$ & $45.0$ \\
    \instance{m21} & $1702.3$ & $7000.0$ & $656.0$ & $2000.0$ & $11901.0$ & $62.7$ & $2281.0$ & $82.5$ \\
    \instance{m22} & $7780.0$ & $8400.0$ & $660.5$ & $986.0$ & $3718.0$ & $296.0$ & $2155.0$ & $15.5$ \\
    \instance{m23} & $6728.5$ & $5400.0$ & $418.6$ & $2040.0$ & $3989.0$ & $169.5$ & $355.1$ & $142.0$ \\
    \instance{m24} & $2813.1$ & $20230.0$ & $1387.6$ & $1616.0$ & $5340.0$ & $88.5$ & $34.0$ & $245.0$ \\
    \instance{m25} & $2364.7$ & $18700.0$ & $396.6$ & $1284.0$ & $11161.0$ & $26.7$ & $344.9$ & $15.2$ \\
    \instance{m26} & $4005.0$ & $48900.0$ & $2601.1$ & $2576.7$ & $12318.9$ & $22.1$ & $2920.0$ & $185.6$ \\
    \instance{m27} & $12530.5$ & $4995.0$ & $2061.0$ & $2160.0$ & $6144.0$ & $422.5$ & $7799.0$ & $17.2$ \\
    \instance{m28} & $23030.0$ & $7875.0$ & $3179.5$ & $4500.0$ & $12766.0$ & $33.9$ & $2673.0$ & $359.0$ \\
    \instance{m29} & $3321.7$ & $34125.0$ & $504.5$ & $3425.0$ & $5827.5$ & $74.0$ & $1392.0$ & $138.0$ \\
    \instance{m30} & $4058.0$ & $9022.2$ & $1425.1$ & $5927.8$ & $11041.1$ & $587.8$ & $428.9$ & $490.0$ \\

\midrule

% python3 table_soft.py default small01 small02 small03 small04 small05 small06 small07 small08 small09
    \instance{small01} & $53.0$ & $0.0$ & $0.0$ & $50.0$ & $130.0$ & $0.0$ & $9.0$ & $0.0$ \\
    \instance{small02} &  &  &  &  &  &  &  &  \\
    \instance{small03} & $380.0$ & $3000.0$ & $30.0$ & $250.0$ & $75.0$ & $0.0$ & $280.0$ & $0.0$ \\
    \instance{small04} & $355.0$ & $3600.0$ & $1.0$ & $240.0$ & $975.0$ & $0.0$ & $71.0$ & $0.0$ \\
    \instance{small05} &  &  &  &  &  &  &  &  \\
    \instance{small06} &  &  &  &  &  &  &  &  \\
    \instance{small07} & $879.5$ & $2000.0$ & $5.5$ & $550.0$ & $480.0$ & $0.0$ & $86.0$ & $0.0$ \\
    \instance{small08} & $890.0$ & $4500.0$ & $10.0$ & $450.0$ & $115.0$ & $0.0$ & $610.0$ & $0.0$ \\
    \instance{small09} & $1121.0$ & $2800.0$ & $65.0$ & $450.0$ & $1156.5$ & $63.5$ & $793.5$ & $0.0$ \\

\midrule

% python3 table_soft.py default l35_1 l35_2 l35_3 l35_4 l35_5 l35_6 l42_1 l42_2 l42_3 l42_4 l42_5 l42_6 l49_1 l49_2 l49_3 l49_4 l49_5 l49_6 l56_1 l56_2 l56_3 l56_4 l56_5 l56_6
    \instance{l35\_1} & $2333.0$ & $1250.0$ & $27.5$ & $560.0$ & $3580.0$ & $40.2$ & $144.6$ & $0.0$ \\
    \instance{l35\_2} & $645.5$ & $27300.0$ & $3.3$ & $390.0$ & $2465.0$ & $4.7$ & $23.9$ & $0.0$ \\
    \instance{l35\_3} & $2632.8$ & $9500.0$ & $1144.0$ & $1656.0$ & $8253.0$ & $1321.0$ & $3905.0$ & $163.0$ \\
    \instance{l35\_4} & $4534.2$ & $2610.0$ & $1876.0$ & $3267.0$ & $17002.0$ & $179.4$ & $2971.0$ & $268.0$ \\
    \instance{l35\_5} & $2453.2$ & $150000.0$ & $3088.0$ & $1500.0$ & $32740.0$ & $321.5$ & $4282.0$ & $5.5$ \\
    \instance{l35\_6} &  &  &  &  &  &  &  &  \\
    \instance{l42\_1} & $910.1$ & $5350.0$ & $1451.0$ & $1302.0$ & $9413.0$ & $48.3$ & $2238.0$ & $0.0$ \\
    \instance{l42\_2} & $1979.6$ & $0.0$ & $2516.1$ & $2120.0$ & $4150.0$ & $86.7$ & $772.0$ & $33.0$ \\
    \instance{l42\_3} & $3233.5$ & $34360.0$ & $2707.0$ & $934.0$ & $37027.0$ & $692.0$ & $8776.5$ & $11.2$ \\
    \instance{l42\_4} & $2364.7$ & $118485.0$ & $1292.9$ & $700.0$ & $6503.5$ & $13.7$ & $298.0$ & $10.0$ \\
    \instance{l42\_5} & $22263.5$ & $76040.0$ & $5581.0$ & $5056.0$ & $30140.0$ & $968.0$ & $4093.0$ & $143.5$ \\
    \instance{l42\_6} & $22708.0$ & $73725.0$ & $3619.5$ & $1057.0$ & $4933.5$ & $204.5$ & $668.5$ & $208.0$ \\
    \instance{l49\_1} & $829.5$ & $15250.0$ & $306.0$ & $1443.0$ & $5519.0$ & $45.5$ & $1726.0$ & $0.0$ \\
    \instance{l49\_2} & $2008.2$ & $54000.0$ & $707.0$ & $762.0$ & $28424.0$ & $902.0$ & $4091.0$ & $108.0$ \\
    \instance{l49\_3} & $19614.5$ & $22000.0$ & $7220.5$ & $2020.0$ & $28290.0$ & $2345.0$ & $783.2$ & $77.0$ \\
    \instance{l49\_4} & $17011.0$ & $62880.0$ & $4157.5$ & $959.0$ & $28617.5$ & $214.6$ & $771.0$ & $14.5$ \\
    \instance{l49\_5} & $8815.5$ & $23400.0$ & $15020.0$ & $5175.0$ & $30585.0$ & $550.0$ & $3550.0$ & $227.5$ \\
    \instance{l49\_6} & $6239.0$ & $84700.0$ & $11455.0$ & $2945.0$ & $20300.0$ & $2837.5$ & $1371.0$ & $181.2$ \\
    \instance{l56\_1} & $8361.0$ & $4000.0$ & $1385.0$ & $1480.0$ & $16370.0$ & $59.2$ & $3405.5$ & $4.0$ \\
    \instance{l56\_2} & $2388.4$ & $31066.7$ & $571.9$ & $2756.7$ & $20274.4$ & $206.2$ & $872.8$ & $140.0$ \\
    \instance{l56\_3} & $4552.7$ & $73605.0$ & $4793.0$ & $2324.0$ & $26105.0$ & $179.6$ & $5545.0$ & $281.0$ \\
    \instance{l56\_4} & $18778.5$ & $172840.0$ & $3950.5$ & $2110.0$ & $14980.5$ & $98.2$ & $3708.0$ & $148.0$ \\
    \instance{l56\_5} & $4860.3$ & $82020.0$ & $7693.0$ & $2388.0$ & $6999.0$ & $1241.5$ & $5568.0$ & $20.9$ \\
    \instance{l56\_6} &  &  &  &  &  &  &  &  \\

  \end{longtable}
\end{center}

\begin{center}
  \setlength{\tabcolsep}{2pt}
  \footnotesize
  \begin{longtable}{%
    p{0.09\textwidth}%
    R{0.08\textwidth}%
    R{0.10\textwidth}%
    R{0.08\textwidth}%
    R{0.08\textwidth}%
    R{0.08\textwidth}%
    R{0.07\textwidth}%
    R{0.08\textwidth}%
    R{0.09\textwidth}%
    R{0.09\textwidth}%
  }
    \caption{%
      Detailed average costs per instance and soft constraint for the 1\,h runs (CoC = Continuity of care, Unsch.\ = unscheduled optional patients, Exc.\ WL = excess workload, Open OTs = opened operating theaters, Delay = delayed patients, Age Mix = age mix in rooms, Skill match = underqualified nurses, surgeon transfer between operating theaters).
      Empty cells indicate that no solution was found.
    }%
    \label{table_results_details_long}\\

\toprule
Instance & CoC & Unsch. & Exc. WL & Open OTs & Delay & Age Mix & Skill Match & Surgeon Transfer \\ 
\midrule
\endfirsthead

\multicolumn{4}{c}{{Table \thetable\ (continued)}}\\
\toprule
Instance & CoC & Unsched. & Excess. WL & Open OTs & Delay & Age Mix & Skill Match & Surgeon Transfer \\ 
\midrule
\endhead

\bottomrule
\endfoot

% python3 table_soft.py long i01 i02 i03 i04 i05 i06 i07 i08 i09 i10 i11 i12 i13 i14 i15 i16 i17 i18 i19 i20 i21 i22 i23 i24 i25 i26 i27 i28 i29 i30
    \instance{i01} & $126.8$ & $2800.0$ & $17.0$ & $240.0$ & $470.0$ & $12.0$ & $219.0$ & $0.0$ \\
    \instance{i02} & $247.0$ & $0.0$ & $30.0$ & $240.0$ & $580.0$ & $30.0$ & $250.0$ & $0.0$ \\
    \instance{i03} & $651.0$ & $7600.0$ & $8.0$ & $400.0$ & $1725.0$ & $20.0$ & $130.5$ & $0.0$ \\
    \instance{i04} & $348.2$ & $0.0$ & $80.5$ & $280.0$ & $946.5$ & $43.0$ & $199.1$ & $0.0$ \\
    \instance{i05} & $348.6$ & $8500.0$ & $3.0$ & $480.0$ & $3427.0$ & $6.5$ & $45.0$ & $0.0$ \\
    \instance{i06} & $234.0$ & $9900.0$ & $0.0$ & $240.0$ & $250.0$ & $12.0$ & $43.4$ & $0.0$ \\
    \instance{i07} & $2403.5$ & $0.0$ & $194.0$ & $344.0$ & $1442.0$ & $71.0$ & $913.0$ & $2.8$ \\
    \instance{i08} & $1123.0$ & $200.0$ & $901.8$ & $1712.0$ & $2192.0$ & $130.0$ & $10.0$ & $71.0$ \\
    \instance{i09} & $2389.0$ & $0.0$ & $185.9$ & $416.0$ & $3453.0$ & $200.5$ & $173.2$ & $42.0$ \\
    \instance{i10} & $2996.0$ & $12780.0$ & $107.0$ & $460.0$ & $2732.0$ & $101.0$ & $1704.5$ & $0.0$ \\
    \instance{i11} & $324.2$ & $24500.0$ & $0.0$ & $570.0$ & $485.0$ & $2.0$ & $74.5$ & $0.0$ \\
    \instance{i12} & $3034.5$ & $5280.0$ & $785.5$ & $819.0$ & $1095.0$ & $12.9$ & $1828.0$ & $18.5$ \\
    \instance{i13} & $2653.5$ & $9500.0$ & $230.9$ & $273.0$ & $6990.0$ & $224.5$ & $263.2$ & $27.0$ \\
    \instance{i14} & $3563.5$ & $3850.0$ & $72.5$ & $1100.0$ & $845.0$ & $17.8$ & $354.6$ & $7.0$ \\
    \instance{i15} & $4375.0$ & $2450.0$ & $215.0$ & $332.0$ & $4610.0$ & $33.5$ & $1083.0$ & $29.5$ \\
    \instance{i16} & $860.3$ & $4140.0$ & $875.8$ & $1368.0$ & $2883.0$ & $64.8$ & $346.0$ & $30.0$ \\
    \instance{i17} & $12124.5$ & $13500.0$ & $2216.0$ & $2040.0$ & $10290.0$ & $629.0$ & $7470.0$ & $88.0$ \\
    \instance{i18} & $790.2$ & $33300.0$ & $0.0$ & $750.0$ & $2654.0$ & $6.5$ & $129.6$ & $2.0$ \\
    \instance{i19} & $11261.5$ & $16500.0$ & $2096.5$ & $2025.0$ & $12889.5$ & $104.8$ & $3069.0$ & $206.0$ \\
    \instance{i20} & $937.3$ & $21300.0$ & $322.0$ & $1372.0$ & $3408.0$ & $47.2$ & $2151.0$ & $5.7$ \\
    \instance{i21} & $8853.0$ & $5225.0$ & $872.5$ & $2775.0$ & $6151.5$ & $174.4$ & $3078.5$ & $20.0$ \\
    \instance{i22} & $13262.5$ & $26010.0$ & $1019.0$ & $1654.0$ & $6393.5$ & $191.6$ & $1540.9$ & $25.0$ \\
    \instance{i23} & $9818.5$ & $19820.0$ & $1529.0$ & $685.0$ & $7396.5$ & $63.9$ & $178.0$ & $6.2$ \\
    \instance{i24} & $2493.6$ & $28350.0$ & $63.6$ & $656.0$ & $1752.5$ & $26.4$ & $422.0$ & $134.0$ \\
    \instance{i25} & $5710.5$ & $2460.0$ & $627.4$ & $1036.0$ & $2012.0$ & $196.5$ & $454.5$ & $75.0$ \\
    \instance{i26} & $17789.0$ & $32200.0$ & $2382.0$ & $857.0$ & $14822.0$ & $230.7$ & $1511.2$ & $98.5$ \\
    \instance{i27} & $4541.0$ & $19900.0$ & $5066.0$ & $3216.0$ & $14293.0$ & $250.6$ & $11507.0$ & $255.0$ \\
    \instance{i28} & $5941.5$ & $63900.0$ & $933.3$ & $2595.0$ & $2158.0$ & $8.1$ & $361.5$ & $48.5$ \\
    \instance{i29} & $1824.2$ & $4200.0$ & $324.0$ & $670.0$ & $4595.0$ & $531.5$ & $683.6$ & $107.0$ \\
    \instance{i30} & $8306.5$ & $16760.0$ & $3003.4$ & $2700.0$ & $6400.5$ & $720.5$ & $194.0$ & $20.0$ \\

\midrule

% python3 table_soft.py long m01 m02 m03 m04 m05 m06 m07 m08 m09 m10 m11 m12 m13 m14 m15 m16 m17 m18 m19 m20 m21 m22 m23 m24 m25 m26 m27 m28 m29 m30
    \instance{m01} & $198.4$ & $5500.0$ & $0.0$ & $440.0$ & $1425.0$ & $0.0$ & $131.6$ & $0.0$ \\
    \instance{m02} & $238.1$ & $10200.0$ & $3.0$ & $700.0$ & $1020.0$ & $4.4$ & $73.5$ & $0.0$ \\
    \instance{m03} & $827.5$ & $5400.0$ & $37.5$ & $80.0$ & $335.0$ & $0.0$ & $76.0$ & $0.0$ \\
    \instance{m04} & $208.2$ & $7350.0$ & $77.7$ & $480.0$ & $1840.0$ & $80.5$ & $1.0$ & $0.0$ \\
    \instance{m05} & $340.9$ & $9600.0$ & $170.2$ & $750.0$ & $951.0$ & $25.0$ & $157.0$ & $0.0$ \\
    \instance{m06} & $2989.5$ & $20050.0$ & $13.0$ & $820.0$ & $3703.5$ & $22.5$ & $581.0$ & $0.0$ \\
    \instance{m07} & $364.6$ & $5700.0$ & $0.0$ & $510.0$ & $1370.0$ & $30.0$ & $163.1$ & $0.0$ \\
    \instance{m08} & $4441.0$ & $8425.0$ & $57.0$ & $480.0$ & $1127.5$ & $25.5$ & $468.5$ & $40.0$ \\
    \instance{m09} & $1213.1$ & $16000.0$ & $1254.3$ & $1170.0$ & $13098.0$ & $6.7$ & $215.0$ & $10.0$ \\
    \instance{m10} & $2606.5$ & $19800.0$ & $200.6$ & $1000.0$ & $1965.0$ & $32.0$ & $496.8$ & $0.0$ \\
    \instance{m11} & $4858.0$ & $17610.0$ & $562.3$ & $2200.0$ & $9730.5$ & $6.0$ & $53.0$ & $35.0$ \\
    \instance{m12} & $4667.5$ & $2655.0$ & $448.1$ & $476.0$ & $2820.0$ & $498.5$ & $623.4$ & $32.5$ \\
    \instance{m13} & $1194.2$ & $17000.0$ & $33.0$ & $483.0$ & $12300.0$ & $45.4$ & $234.7$ & $44.0$ \\
    \instance{m14} & $1047.6$ & $9100.0$ & $3.0$ & $630.0$ & $1843.0$ & $7.0$ & $692.5$ & $34.5$ \\
    \instance{m15} & $1851.5$ & $6000.0$ & $2419.9$ & $1416.0$ & $16311.0$ & $896.5$ & $74.0$ & $152.0$ \\
    \instance{m16} & $5449.5$ & $5250.0$ & $2504.5$ & $1684.0$ & $1766.0$ & $30.2$ & $1690.0$ & $64.5$ \\
    \instance{m17} & $1054.9$ & $17000.0$ & $1309.6$ & $660.0$ & $2264.0$ & $12.2$ & $308.0$ & $6.0$ \\
    \instance{m18} & $1385.4$ & $2250.0$ & $1505.6$ & $428.0$ & $2435.5$ & $425.5$ & $20.0$ & $57.0$ \\
    \instance{m19} & $1085.4$ & $11970.0$ & $179.0$ & $1203.0$ & $7310.0$ & $371.0$ & $200.5$ & $76.0$ \\
    \instance{m20} & $1381.4$ & $1200.0$ & $635.6$ & $420.0$ & $1493.5$ & $247.5$ & $98.5$ & $50.0$ \\
    \instance{m21} & $1378.9$ & $6500.0$ & $325.0$ & $2000.0$ & $12106.5$ & $68.9$ & $1946.0$ & $108.0$ \\
    \instance{m22} & $7335.0$ & $8560.0$ & $746.0$ & $1004.0$ & $3719.0$ & $378.0$ & $2028.0$ & $15.0$ \\
    \instance{m23} & $6705.0$ & $5430.0$ & $441.8$ & $2032.0$ & $3976.0$ & $242.0$ & $329.0$ & $156.0$ \\
    \instance{m24} & $2543.8$ & $19985.0$ & $1433.8$ & $1616.0$ & $5331.0$ & $204.5$ & $26.0$ & $243.0$ \\
    \instance{m25} & $2353.1$ & $19220.0$ & $278.7$ & $1280.0$ & $10951.0$ & $24.1$ & $375.5$ & $15.7$ \\
    \instance{m26} & $4083.5$ & $48510.0$ & $2518.0$ & $2577.0$ & $12079.0$ & $87.4$ & $3516.0$ & $202.0$ \\
    \instance{m27} & $12005.5$ & $5085.0$ & $2116.0$ & $2132.0$ & $6123.0$ & $555.0$ & $7508.5$ & $17.5$ \\
    \instance{m28} & $22315.7$ & $7750.0$ & $3037.1$ & $4491.4$ & $12707.1$ & $44.0$ & $2622.9$ & $341.4$ \\
    \instance{m29} & $2691.8$ & $41535.0$ & $376.5$ & $3040.0$ & $9468.0$ & $197.5$ & $1154.0$ & $116.0$ \\
    \instance{m30} & $4027.8$ & $8880.0$ & $1485.4$ & $5930.0$ & $11182.0$ & $619.0$ & $430.0$ & $428.0$ \\

\midrule

% python3 table_soft.py long small01 small02 small03 small04 small05 small06 small07 small08 small09
    \instance{small01} & $53.0$ & $0.0$ & $0.0$ & $50.0$ & $130.0$ & $0.0$ & $9.0$ & $0.0$ \\
    \instance{small02} &  &  &  &  &  &  &  &  \\
    \instance{small03} & $380.0$ & $3000.0$ & $30.0$ & $250.0$ & $75.0$ & $0.0$ & $280.0$ & $0.0$ \\
    \instance{small04} & $355.0$ & $3600.0$ & $1.0$ & $240.0$ & $975.0$ & $0.0$ & $71.0$ & $0.0$ \\
    \instance{small05} &  &  &  &  &  &  &  &  \\
    \instance{small06} &  &  &  &  &  &  &  &  \\
    \instance{small07} & $878.5$ & $2000.0$ & $6.5$ & $550.0$ & $480.0$ & $0.0$ & $86.0$ & $0.0$ \\
    \instance{small08} & $890.0$ & $4500.0$ & $10.0$ & $450.0$ & $115.0$ & $0.0$ & $610.0$ & $0.0$ \\
    \instance{small09} & $1122.0$ & $2800.0$ & $70.0$ & $450.0$ & $1155.0$ & $60.0$ & $793.0$ & $0.0$ \\

\midrule

% python3 table_soft.py long l35_1 l35_2 l35_3 l35_4 l35_5 l35_6 l42_1 l42_2 l42_3 l42_4 l42_5 l42_6 l49_1 l49_2 l49_3 l49_4 l49_5 l49_6 l56_1 l56_2 l56_3 l56_4 l56_5 l56_6
    \instance{l35\_1} & $2322.0$ & $1250.0$ & $14.5$ & $560.0$ & $3580.0$ & $40.0$ & $149.0$ & $0.0$ \\
    \instance{l35\_2} & $627.6$ & $27300.0$ & $2.3$ & $390.0$ & $2358.5$ & $3.8$ & $24.1$ & $0.0$ \\
    \instance{l35\_3} & $2656.0$ & $9400.0$ & $1146.0$ & $1668.0$ & $8056.5$ & $1422.0$ & $3971.0$ & $153.0$ \\
    \instance{l35\_4} & $4447.0$ & $2385.0$ & $2296.0$ & $3222.0$ & $16627.5$ & $213.4$ & $2894.5$ & $262.0$ \\
    \instance{l35\_5} & $4732.6$ & $26900.0$ & $10588.5$ & $2382.0$ & $31740.0$ & $3119.0$ & $9030.0$ & $32.3$ \\
    \instance{l35\_6} &  &  &  &  &  &  &  &  \\
    \instance{l42\_1} & $911.0$ & $5200.0$ & $1479.5$ & $1248.0$ & $9279.5$ & $70.0$ & $2174.0$ & $0.0$ \\
    \instance{l42\_2} & $1976.4$ & $0.0$ & $2544.5$ & $2128.0$ & $4089.0$ & $97.0$ & $779.0$ & $31.0$ \\
    \instance{l42\_3} & $2700.1$ & $33520.0$ & $1000.0$ & $942.0$ & $37152.0$ & $799.0$ & $7390.0$ & $13.0$ \\
    \instance{l42\_4} & $2276.0$ & $117900.0$ & $1449.9$ & $700.0$ & $6288.5$ & $30.8$ & $253.0$ & $10.0$ \\
    \instance{l42\_5} & $21905.0$ & $76120.0$ & $4629.0$ & $5080.0$ & $29938.0$ & $994.5$ & $4016.0$ & $140.5$ \\
    \instance{l42\_6} & $21649.5$ & $72900.0$ & $3748.8$ & $1051.0$ & $4681.0$ & $458.5$ & $690.5$ & $208.0$ \\
    \instance{l49\_1} & $844.9$ & $15125.0$ & $276.5$ & $1455.0$ & $5512.0$ & $29.0$ & $1728.0$ & $0.0$ \\
    \instance{l49\_2} & $1675.7$ & $53730.0$ & $419.0$ & $761.0$ & $28556.0$ & $902.0$ & $3716.0$ & $110.0$ \\
    \instance{l49\_3} & $19094.5$ & $21440.0$ & $6334.0$ & $2008.0$ & $28780.5$ & $2497.0$ & $783.1$ & $94.0$ \\
    \instance{l49\_4} & $13506.0$ & $62340.0$ & $4544.3$ & $960.0$ & $28474.0$ & $255.7$ & $1116.0$ & $14.1$ \\
    \instance{l49\_5} & $8681.2$ & $24360.0$ & $12286.0$ & $5106.0$ & $31116.0$ & $522.4$ & $3688.0$ & $227.0$ \\
    \instance{l49\_6} & $6060.8$ & $86875.0$ & $9227.5$ & $2892.5$ & $19563.8$ & $2304.4$ & $1312.6$ & $157.5$ \\
    \instance{l56\_1} & $7979.0$ & $3500.0$ & $1187.0$ & $1500.0$ & $16140.0$ & $70.4$ & $3630.5$ & $3.1$ \\
    \instance{l56\_2} & $2121.0$ & $30210.0$ & $470.2$ & $2760.0$ & $20456.0$ & $272.4$ & $856.6$ & $159.0$ \\
    \instance{l56\_3} & $4696.9$ & $72765.0$ & $3450.0$ & $2304.0$ & $25535.0$ & $220.2$ & $7432.5$ & $285.0$ \\
    \instance{l56\_4} & $22260.5$ & $134160.0$ & $6781.0$ & $2312.0$ & $10130.5$ & $327.6$ & $5766.0$ & $170.0$ \\
    \instance{l56\_5} & $4901.9$ & $82065.0$ & $6557.0$ & $2362.0$ & $7158.0$ & $1248.0$ & $5051.0$ & $21.7$ \\
    \instance{l56\_6} & $6804.2$ & $153650.0$ & $4170.0$ & $8491.7$ & $60380.0$ & $913.3$ & $1699.7$ & $141.7$ \\

  \end{longtable}
\end{center}

\pagebreak[3]

\Cref{table_results_extra2} presents the results of \cref{table_results_extra1} for the non-competition instances, which are discussed in \cref{sec_experiments_results_sa,sec_experiments_results_rejections}.

\begin{center}
  \setlength{\tabcolsep}{2pt}
  \footnotesize%
  \begin{longtable}{%
p{0.08\textwidth}%
R{0.08\textwidth}%
R{0.09\textwidth}%
R{0.11\textwidth}R{0.11\textwidth}R{0.07\textwidth}%
R{0.07\textwidth}%
R{0.11\textwidth}%
R{0.07\textwidth}%
}%
    \caption{%
      Results for our default algorithm and the variant described in \cref{sec_rejection} for \SI{10}{\minute}, each \edit{run} 10 times per non-competition instance.
      See \cref{table_results_extra1} for the competition instances.
      Similar to \cref{table_results_quality}, lower bounds and best known solutions are shown.
      For each of the algorithms, the (averaged) objective value (column obj) of the best found solution and the number of produced admission solutions are reported.
      A number in brackets after the objective value indicates the number of runs in which no feasible solution was found.
      In this case, the averages are computed over the remaining runs.
      In addition, column nurse\,$\searrow$ shows the relative improvement of the nurse costs (sum of continuity of care, excess workload and skill match) from the end of Phase~2 to the end of Phase~3.
    }
    \label{table_results_extra2} \\
\toprule
    Instance & Lower & Best sol & \multicolumn{3}{c}{Default \SI{10}{\minute}} & $\searrow$ & \multicolumn{2}{c}{Rejection \SI{10}{\minute}} \\
\cmidrule(r){4-6} \cmidrule(r){8-9} 
          & bound & Obj. & Nurse\,$\searrow$  & Obj.\,[fails] & \#sols & [\%] & Obj.\,[fails] & \#sols \\
\midrule
\endfirsthead

\multicolumn{4}{c}{{Table \thetable\ (continued)}}\\
\toprule
    Instance & Lower & Best sol & \multicolumn{3}{c}{Default \SI{10}{\minute}} & $\searrow$ & \multicolumn{2}{c}{Rejection \SI{10}{\minute}} \\
\cmidrule(r){4-6} \cmidrule(r){8-9} 
          & bound & Obj. & Nurse\,$\searrow$  & Obj.\,[fails] & \#sols & [\%] & Obj.\,[fails] & \#sols \\
\midrule
\endhead

\bottomrule
\endfoot

% python3 table_extras.py small01 small02 small03 small04 small05 small06 small07 small08 small09
    \instance{small01} & \num{233} & $\infty$      & $0.0\,\%$ & \num{242}\phantom{\,[0]} & \num{12.0}    & $0.0$ & \num{242}\phantom{\,[0]} & \num{4859.7} \\
    \instance{small02} & \num{690} & $\infty$      &  & $\infty$\phantom{{\,[0]}} & \num{12.7}    &  & \num{2136}\phantom{\,[0]} & \num{50.1} \\
    \instance{small03} & \num{3860} & $\infty$      & $1.4\,\%$ & \num{4015}\phantom{\,[0]} & \num{11.0}    & $0.0$ & \num{4015}\phantom{\,[0]} & \num{449.0} \\
    \instance{small04} & \num{1992} & $\infty$      & $0.5\,\%$ & \num{5242}\phantom{\,[0]} & \num{17.0}    & $\better{27.1}$ & \num{3822}\phantom{\,[0]} & \num{2010.9} \\
    \instance{small05} & \num{3215} & $\infty$      &  & $\infty$\phantom{{\,[0]}} & \num{2.0}    &  & $\infty$\phantom{{\,[0]}} & \num{12.1} \\
    \instance{small06} & \num{2534} & $\infty$      &  & $\infty$\phantom{{\,[0]}} & \num{26.8}    &  & \num{4739}\phantom{\,[0]} & \num{47.0} \\
    \instance{small07} & \num{2797} & $\infty$      & $7.5\,\%$ & \num{4001}\phantom{\,[0]} & \num{57.0}    & $\better{12.1}$ & \num{3518}\phantom{\,[0]} & \num{1922.7} \\
    \instance{small08} & \num{6085} & $\infty$      & $1.0\,\%$ & \num{6575}\phantom{\,[0]} & \num{19.0}    & $\better{0.3}$ & \num{6553}\phantom{\,[0]} & \num{1140.3} \\
    \instance{small09} & \num{5285} & $\infty$      & $8.9\,\%$ & \num{6450}\phantom{\,[0]} & \num{111.0}    & $\better{0.7}$ & \num{6404}\phantom{\,[0]} & \num{1465.2} \\

\midrule

% python3 table_extras.py l35_1 l35_2 l35_3 l35_4 l35_5 l35_6 l42_1 l42_2 l42_3 l42_4 l42_5 l42_6 l49_1 l49_2 l49_3 l49_4 l49_5 l49_6 l56_1 l56_2 l56_3 l56_4 l56_5 l56_6
    \instance{l35\_1} & \num{6621} & $\infty$      & $13.9\,\%$ & \num{7935}\phantom{\,[0]} & \num{42.5}    & $\worse{-4.6}$ & \num{8297}\phantom{\,[0]} & \num{341.9} \\
    \instance{l35\_2} & \num{30517} & $\infty$      & $11.9\,\%$ & \num{30832}\phantom{\,[0]} & \num{41.7}    & $\better{0.0}$ & \num{30822}\phantom{\,[0]} & \num{186.1} \\
    \instance{l35\_3} & \num{19098} & $\infty$      & $2.7\,\%$ & \num{28575}\phantom{\,[0]} & \num{94.6}    & $\worse{-1.4}$ & \num{28970}\phantom{\,[0]} & \num{239.0} \\
    \instance{l35\_4} & \num{23060} & $\infty$      & $3.6\,\%$ & \num{32708}\phantom{\,[0]} & \num{63.8}    & $\worse{-19.1}$ & \num{38941}\phantom{\,[0]} & \num{101.9} \\
    \instance{l35\_5} & \num{60567} & $\infty$      & $6.3\,\%$ & \num{194390}\phantom{\,[0]} & \num{86.6}    & $\better{52.4}$ & \num{92523}\phantom{\,[0]} & \num{174.3} \\
    \instance{l35\_6} & \num{51445} & $\infty$      &  & $\infty$\phantom{{\,[0]}} & \num{189.5}    &  & \num{62560}\,[6] & \num{225.2} \\
    \instance{l42\_1} & \num{13100} & $\infty$      & $6.1\,\%$ & \num{20712}\phantom{\,[0]} & \num{100.2}    & $\better{1.1}$ & \num{20482}\phantom{\,[0]} & \num{205.8} \\
    \instance{l42\_2} & \num{8095} & $\infty$      & $27.6\,\%$ & \num{11657}\phantom{\,[0]} & \num{112.4}    & $\worse{-0.0}$ & \num{11659}\phantom{\,[0]} & \num{206.4} \\
    \instance{l42\_3} & \num{70554} & $\infty$      & $2.4\,\%$ & \num{87741}\phantom{\,[0]} & \num{130.0}    & $\worse{-6.1}$ & \num{93137}\phantom{\,[0]} & \num{145.6} \\
    \instance{l42\_4} & \num{126131} & $\infty$      & $76.8\,\%$ & \num{129668}\phantom{\,[0]} & \num{55.5}    & $\worse{-0.7}$ & \num{130574}\phantom{\,[0]} & \num{72.0} \\
    \instance{l42\_5} & \num{120305} & $\infty$      & $29.9\,\%$ & \num{144285}\phantom{\,[0]} & \num{78.3}    & $\worse{-1.5}$ & \num{146521}\phantom{\,[0]} & \num{178.3} \\
    \instance{l42\_6} & \num{89792} & $\infty$      & $44.9\,\%$ & \num{107124}\phantom{\,[0]} & \num{40.8}    & $\worse{-0.5}$ & \num{107667}\phantom{\,[0]} & \num{22.5} \\
    \instance{l49\_1} & \num{19051} & $\infty$      & $15.3\,\%$ & \num{25119}\phantom{\,[0]} & \num{79.1}    & $\worse{-0.3}$ & \num{25190}\phantom{\,[0]} & \num{165.2} \\
    \instance{l49\_2} & \num{82164} & $\infty$      & $2.2\,\%$ & \num{91002}\phantom{\,[0]} & \num{94.6}    & $\worse{-1.1}$ & \num{91995}\phantom{\,[0]} & \num{250.3} \\
    \instance{l49\_3} & \num{60007} & $\infty$      & $19.5\,\%$ & \num{82350}\phantom{\,[0]} & \num{204.8}    & $\worse{-2.8}$ & \num{84630}\phantom{\,[0]} & \num{509.3} \\
    \instance{l49\_4} & \num{101492} & $\infty$      & $31.9\,\%$ & \num{114625}\phantom{\,[0]} & \num{41.8}    & $\worse{-0.9}$ & \num{115686}\phantom{\,[0]} & \num{102.1} \\
    \instance{l49\_5} & \num{64641} & $\infty$      & $9.1\,\%$ & \num{87323}\,[8] & \num{84.9}    & $\worse{-0.1}$ & \num{87413}\,[7] & \num{136.4} \\
    \instance{l49\_6} & \num{110208} & $\infty$      & $16.5\,\%$ & \num{130029}\,[6] & \num{89.6}    & $\worse{-1.4}$ & \num{131910}\,[4] & \num{163.0} \\
    \instance{l56\_1} & \num{25775} & $\infty$      & $8.5\,\%$ & \num{35065}\phantom{\,[0]} & \num{134.1}    & $\worse{-6.3}$ & \num{37280}\phantom{\,[0]} & \num{202.8} \\
    \instance{l56\_2} & \num{50028} & $\infty$      & $0.2\,\%$ & \num{58277}\,[1] & \num{138.9}    & $\worse{-12.3}$ & \num{65445}\phantom{\,[0]} & \num{116.2} \\
    \instance{l56\_3} & \num{101624} & $\infty$      & $35.0\,\%$ & \num{117385}\phantom{\,[0]} & \num{109.2}    & $\worse{-2.5}$ & \num{120331}\phantom{\,[0]} & \num{79.2} \\
    \instance{l56\_4} & \num{152835} & $\infty$      & $43.3\,\%$ & \num{216614}\phantom{\,[0]} & \num{80.2}    & $\better{2.3}$ & \num{211577}\,[2] & \num{66.8} \\
    \instance{l56\_5} & \num{93126} & $\infty$      & $33.6\,\%$ & \num{110791}\phantom{\,[0]} & \num{57.8}    & $\worse{-0.2}$ & \num{110987}\phantom{\,[0]} & \num{281.0} \\
    \instance{l56\_6} & \num{182912} & $\infty$      &  & $\infty$\phantom{{\,[0]}} & \num{33.5}    &  & $\infty$\phantom{{\,[0]}} & \num{15.7} \\

  \end{longtable}
\end{center}

\end{document}